\documentclass{article}
\usepackage{float}
\usepackage{shapepar}
\usepackage{listings}
\usepackage{amsthm}
\usepackage{mathtools}
\usepackage{graphics}
\usepackage{amssymb}
\usepackage{mathrsfs}
\usepackage{framed}
\usepackage{diagbox}
\usepackage{booktabs}
\usepackage{fancybox}
\usepackage{geometry}
\usepackage{multirow}
\usepackage{enumerate}
\usepackage{caption}
\usepackage{subcaption}
\usepackage{hyperref}
\usepackage{boldline}
\usepackage[table]{xcolor}
\usepackage{slashbox}
\usepackage{tabularray}
\usepackage{vcell}
\usepackage{titlesec}

\usepackage{makecell}
\usepackage{algorithmic}
\usepackage{lscape}
\usepackage{verbatim}
\usepackage{arydshln}
\setcellgapes{5pt}

\usepackage[pagewise]{lineno}

\titleformat{\paragraph}
{\normalfont\normalsize\bfseries}{\theparagraph}{1em}{}
\titlespacing*{\paragraph}
{0pt}{3.25ex plus 1ex minus .2ex}{1.5ex plus .2ex}

\usepackage[ruled,linesnumbered]{algorithm2e}
\usepackage[square,numbers]{natbib}

\newtheorem{thm}{Theorem}[section]

\newcommand \R {\mathbb{R}}
\newcommand \C {\mathbb{C}}

\SetCommentSty{mycommfont}

\usepackage{authblk}

\usepackage{abstract}

\title{Electrical Impedance Tomography: A Fair Comparative Study on Deep Learning and Analytic-based Approaches}
\date{}

\author[$\ddagger *$]{Derick Nganyu Tanyu}
\author[$\dagger *$]{Jianfeng Ning}
\author[$\natural$]{Andreas Hauptmann}
\author[$\S$]{\\Bangti Jin}
\author[$\ddagger$]{Peter Maass}

\affil[$\ddagger$]{\small Centre for Industrial Mathematics (ZeTeM), University of Bremen, Bremen, Germany}
\affil[$\dagger$]{\small School of Mathematics and Statistics, Wuhan University, Wuhan, China}
\affil[$\natural$]{\small Research Unit of Mathematical Sciences, University of Oulu, Oulu, Finland}
\affil[$\S$]{\small Department of Mathematics, The Chinese University of Hong Kong, Shatin, N.T., Hong Kong.}
\affil[*]{Corresponding authors, equal contribution: \href{mailto:derick@uni-bremen.de}{derick@uni-bremen.de} , \href{mailto:ningjf@whu.edu.cn}{ningjf@whu.edu.cn}}

\begin{document}
\maketitle

\begin{abstract}
Electrical Impedance Tomography (EIT) is a powerful imaging technique with diverse applications, e.g., medical diagnosis, industrial monitoring, and environmental studies. The EIT inverse problem is about inferring the internal conductivity distribution of an object from measurements taken on its boundary. It is severely ill-posed, necessitating advanced computational methods for accurate image reconstructions. Recent years have witnessed significant progress, driven by innovations in analytic-based approaches and deep learning.
This review comprehensively explores techniques for solving the EIT inverse problem, focusing on the interplay between contemporary deep learning-based strategies and classical analytic-based methods. Four state-of-the-art deep learning algorithms are rigorously examined, including the deep D-bar method, deep direct sampling method, fully connected U-net, and convolutional neural networks, harnessing the representational capabilities of deep neural networks to reconstruct intricate conductivity distributions. In parallel, two analytic-based methods, i.e., sparsity regularisation and D-bar method, rooted in mathematical formulations and regularisation techniques, are dissected for their strengths and limitations. These methodologies are evaluated through an extensive array of numerical experiments, encompassing diverse scenarios that reflect real-world complexities.
A suite of performance metrics is employed to assess the efficacy of these methods. These metrics collectively provide a nuanced understanding of the methods' ability to capture essential features and delineate complex conductivity patterns.

One novel feature of the study is the incorporation of variable conductivity scenarios, introducing a level of heterogeneity that mimics textured inclusions. This departure from uniform conductivity assumptions mimics realistic scenarios where tissues or materials exhibit spatially varying electrical properties. Exploring how each method responds to such variable conductivity scenarios opens avenues for understanding their robustness and adaptability.
\end{abstract}

\section{Introduction and motivation}   
This paper investigates deep learning concepts for the continuous model of electrical impedance tomography (EIT). EIT is one of the most intensively studied inverse problems, and there already exists a very rich body of literature on various aspects \cite{Borcea:2002,Uhlmann:2009}. 
EIT as an imaging modality is of
considerable practical interest in noninvasive
imaging and non-destructive testing. For example, the reconstruction
can be used for diagnostic purposes in medical applications, e.g.
monitoring of lung function, detection of cancer in the skin and
breast and location of epileptic foci \cite{Ho:04}. Similarly, in geophysics, one uses electrodes on
the surface of the earth or in boreholes to locate resistivity
anomalies, e.g. minerals or contaminated sites, and it is known as
geophysical resistivity tomography in the literature.

Since its first formulation by Calder\'on \cite{Ca:80}, the issue of image reconstruction has received enormous attention, and many reconstruction algorithms have been proposed based on regularised reconstructions, e.g., Sobolev smoothness, total variation and sparsity. Due to the severe ill-posed nature of the inverse problem and the high degree of non-linearity of the forward model, the resolution of the obtained reconstructions has been modest at best. Nonetheless, the last years have witnessed significant improvement in the EIT reconstruction regarding resolution and speed. This impressive progress was primarily driven by recent innovations in deep learning, especially deep neural network architectures, high-quality paired training data, efficient training algorithms (e.g., Adam), and powerful computing facilities, e.g., graphical processing units (GPUs).

This study aims to comprehensively and fairly compare deep learning techniques for solving the EIT inverse problem. This study has several sources of motivation. First, the classical, analytical setting of EIT is severely ill-posed, to such an extent that it allows only rather sketchy reconstructions when employing classical regularisation schemes. Unless one utilises additional \textit{a priori} information, there is no way around the ill-posedness. This has motivated the application of learning concepts in this context. Incorporating additional information in the form of typical data sets and ground truth reconstructions allows constructing an approximation of a data manifold specific to the task at hand. The structures that distinguish these manifolds are typically hard to capture by explicit physical-mathematical models. To some extent, TV- or sparsity-based Tikhonov functionals exploit these features. However, learning the prior distribution from sufficiently large sets of training data potentially offers much greater flexibility than these hand-crafted priors. Second, there already exists a growing and rich body of literature on learned concepts for EIT; see, e.g., the recent survey \cite{khan2019review} and Section \ref{sec:DL} for a detailed description of the state of the art. Nevertheless, most of these works focus on their own approaches, typically showing their superiority compared to somewhat standard and basic analytical methods. In contrast, we aim at a fair and more comprehensive comparison of different learned concepts and include a comparison with two advanced analytical methods (i.e., D-bar and sparsity methods). 

It is worth mentioning that inverse problems pose a particular challenge for learned concepts due to their inherent instability. For example, directly adapting well-established network architectures, which have been successfully applied to computer vision or imaging problems, typically fail for inverse problems, e.g., medical image reconstruction tasks. Hence, such learned concepts for inverse parameter identification problems are most interesting in terms of developing an underlying theory and the performance on practical applications. 
Indeed, the research on learned concepts for inverse problems has exploded over the past years, see e.g. the review \cite{arridge_acta} and the references cited therein for a recent overview of the state of the art.
Arguably, the two most prominent fields of application for inverse problems are PDE-based parameter identification problems and tasks in tomographic image reconstruction. These fields actually overlap, e.g. when it comes to parameter identification problems in PDE-based multi-physics models for imaging. The most common examples in tomography are X-ray tomography (linear) and EIT (non-linear). Hence, one may also regard this study as being prototypical of how deep learning concepts should be evaluated in the context of non-linear PDE inverse problems.

The rest of the paper is organised as follows. In Section \ref{sec:EIT}, we describe the continuum model for EIT, and also two prominent analytic-based approaches for EIT reconstruction, i.e., sparsity and D-bar method. Then, in Section \ref{sec:DL}, we describe four representative deep learning-based approaches for EIT imaging. Finally, in Section \ref{sec:experiment}, we present an extensive array of experiments with a suite of performance metrics to shed insights into the relative merits of the methods.  We conclude with further discussions in Section \ref{sec:discussion}.

\section{Electrical impedance tomography}\label{sec:EIT}
Mathematically speaking, the continuous EIT problem aims at determining a
spatially-varying electrical conductivity $\sigma$ within a bounded
domain $\Omega$ by using measurements of the electrical potential on the
boundary $\partial\Omega$. The basic mathematical model for the forward
problem is the following elliptic PDE:
\begin{equation}
-\mbox{div}(\sigma\nabla u) = 0,\quad \mbox{ in } \Omega,
\label{eqn:eit}
\end{equation}
subject to a Neumann boundary condition $\sigma\frac{\partial
u}{\partial n}=j$ on $\partial\Omega$, which satisfies a compatibility condition $\int_{\partial\Omega} j {\rm d}S=0$.
An EIT experiment consists of applying an electrical current $j$ on
the boundary $\partial\Omega$ and measuring the resulting electrical
potential $\phi = u|_{\partial\Omega}$ on $\partial\Omega$. 
The Neumann to Dirichlet (NtD) operator $\Lambda_{\sigma,N}:j \mapsto \phi$ maps a Neumann boundary condition $j$ to the Dirichlet data $\phi=u|_{\partial \Omega}$ on $\partial \Omega$.
 
In practice, several input currents are injected, and the induced electrical potentials
are measured; see \cite{CheneyIsaacson:1992,jin2011function} for discussions on the choice of
optimal input currents. This data contains information about the underlying NtD map $\Lambda_{\sigma,N}$. The inverse problem is to determine or at least to approximate the true
unknown physical electrical conductivity $\sigma^\dagger$ from a
partial knowledge of the map. This inverse problem was first
formulated by Calder\'{o}n \cite{Ca:80}, who also gave a uniqueness
result for the linearised problem. The mathematical theory of uniqueness
of the inverse problem with the full NtD map $\Lambda_{\sigma,N}$ has received enormous attention, and
many profound theoretical results have been obtained. For an
in-depth overview of uniqueness results, we refer to the monograph
\cite{Is:06} and survey \cite{Uhlmann:2009}. 
  
\subsection{Theoretical background}
This section introduces the mathematical model of the EIT problem
and the discrepancy functional used for reconstructing the
conductivity $\sigma$.  Let $\Omega$ be an open-bounded domain in $\mathbb{R}^d\ (d\geq2)$
with a Lipschitz boundary $\partial\Omega$,
and let $\Lambda_{\sigma,N}$ denote the NtD map of problem \eqref{eqn:eit}. We employ the usual Sobolev space for the  Neumann boundary data $\sigma\frac{\partial u}{\partial
n}=j\in\tilde{H}^{-\frac{1}{2}}(\partial\Omega)$, respectively Dirichlet
boundary condition $u=\phi\in\tilde{H}^\frac{1}{2}(\partial\Omega)$ on
$\partial\Omega$. Throughout, we make use of the space
$\tilde{H}^1(\Omega)$, which is a subspace of the Sobolev space
$H^1(\Omega)$ with vanishing mean on $\partial\Omega$, i.e., 
$\tilde{H}^1(\Omega)=\{v\in H^1(\Omega): \int_{\partial\Omega} v{\rm d}s=0\}$.
The spaces $\tilde{H}^{\frac{1}{2}}(\partial\Omega)$ and
$\tilde{H}^{-\frac{1}{2}}(\partial\Omega)$ are defined similarly. These
spaces are equipped with the usual norms. We normalise the solution of the Neumann problem by
enforcing $\int_{\partial\Omega}u {\rm d}s = 0$, so that there exists a unique
solution $u\in \tilde{H}^1(\Omega)$. We denote the Dirichlet-to-Neumann (DtN) map by $\Lambda_{\sigma,D}$. Then we have $\Lambda_{\sigma,N}=\Lambda_{\sigma,D}^{-1}$, i.e., DtN and NtD maps are inverse to each other. In usual regularised reconstruction, we employ the NtD map $\Lambda_{\sigma,N}$, whereas in the D-bar method, we employ the DtN map $\Lambda_{\sigma,D}$. 

An EIT experiment consists of applying a current $j$ and measuring the
resulting potential $\phi$ on $\partial\Omega$, and it is equivalent to solving a
Neumann forward problem with the physical conductivity $\sigma^\dagger$, i.e.
$\phi = \Lambda_{\sigma,N} j$, on $\partial\Omega$.
In practice, the boundary
potential measurements are collected experimentally, and thus $\phi$
is only an element of the space $L^2(\Gamma)$. 
see e.g. \cite{CEJ:08}.
Note that the continuum model is mostly academic. A more realistic model is the so-called complete electrode model (CEM) for EIT \cite{SomersaloCheney:1992,Hyvonen:2004}, which models contact impedances and localised electrode geometries. The CEM is finite-dimensional by construction, leading to different mathematical challenges and reconstruction methods.

The solvability, uniqueness and smoothness of the continuum model  with
respect to $L^p$ norms can be derived using Meyers'
gradient estimate \cite{Meyers:1963}, as in \cite{RondiSantosa:2001}.

\begin{thm}\label{thm:meyers}
Let $\Omega$ be a bounded Lipschitz domain in
$\mathbb{R}^d\ (d\geq2)$. Assume that $\sigma\in L^\infty(\Omega)$
satisfies $\lambda<\sigma<\lambda^{-1}$ for some fixed
$\lambda\in(0,1)$. For $f\in (L^q(\Omega))^d$ and $h\in
L^q(\Omega)$, let $u \in H^1(\Omega)$ be a weak solution of
\begin{equation*}
-\mathrm{div}(\sigma\nabla u)=-\mathrm{div}(f) + h\quad \mathrm{in
}\ \Omega.
\end{equation*}
Then, there exists a constant $Q\in(2,+\infty)$ depending on
$\lambda$ and $d$ only, $Q\rightarrow2$ as $\lambda\rightarrow 0$
and $Q\rightarrow\infty$ as $\lambda \rightarrow 1$, such that  for
any $2 < q< Q$, we obtain $u\in W_\mathrm{loc}^{1,q}(\Omega)$ and for
any $\Omega_1\subset\subset\Omega$
\begin{equation*}
\|u\|_{W^{1,q}(\Omega_1)}\leq
C(\|u\|_{H^1(\Omega)}+\|f\|_{L^q(\Omega)} +\|h\|_{L^q(\Omega)}),
\end{equation*}
where the constant $C$ depends on $\lambda$, $d$, $q$, $\Omega_1$
and $\Omega$.
\end{thm}

In Theorem \ref{thm:meyers}, the boundary condition for the problem
can be general. Its effect enters the $W^{1,q}$-estimate through the
term $\|u\|_{H^1(\Omega)}$. In addition, no regularity has been
assumed on $\sigma$. Generally, a precise estimate
of the constant $Q(\lambda,d)$ is missing, but in the 2D case, a fairly sharp estimate of $Q(\lambda,d)$ was derived in \cite{AstalaFaracoSzekelyhidi:2008}.  

\subsection{Conventional EIT reconstruction algorithms} 

EIT suffers from a high degree of
non-linearity and severe ill-posedness, as typical of many PDE inverse problems with boundary data. However, its potential
applications have sparked much interest in designing effective numerical
techniques for its efficient solution. Numerous numerical
methods have been proposed in the literature; see \cite[Section 7]{Borcea:2002} for an overview (up to 2002).
These methods can roughly be divided into two groups: regularised reconstruction and direct methods. Below, we give a brief categorisation of conventional reconstruction schemes.

The methods in the first group are of variational type, i.e., based on minimising a certain discrepancy functional.  Commonly the discrepancy $J$ is the standard least-squares fitting, i.e., the squared $L^2(\partial\Omega)$ norm of the difference
between the electrical potential due to the applied current $j$ and
the measured potential $\phi$:
\begin{equation*}
  J(\sigma)=\tfrac{1}{2}\|\Lambda_{\sigma,N}j-\phi^\delta\|_{L_2(\partial\Omega)}^2,
\end{equation*}
for one single measurement $(j,\phi^\delta)$. One early approach of this type is given in \cite{CheneyIsaacson:1990}, which applies one step of a Newton method with a constant conductivity as the
initial guess. Due to the severe ill-posedness of the
problem, regularisation is beneficial for obtaining reconstructions with improved resolution \cite{EnglHankeNeubauer:1996,HofmannKaltenbacher:2012,ItoJinL2015}. Commonly used penalties include Sobolev smoothness \cite{LukaschewitschMaassPidcock:2003,jin2012analysis} for a smooth conductivity distribution, total variation \cite{HinzeKaltenbacher:2018}, Mumford-Shah functional \cite{RondiSantosa:2001}, level set method \cite{ChungChanTai:2005} for recovering piecewise constant conductivity, sparsity \cite{gehre2012sparsity,jin2017sparsity,Jin_2012} for recovering small inclusions (relative to the background).  The combined functional is given by 
\begin{equation*}
    \Psi(\sigma) = J(\sigma) + \alpha R(\sigma),
\end{equation*}
where $R(\sigma)$ denotes the penalty, and $\alpha>0$ is the penalty weight.
The functional $\Psi(\sigma)$ is then minimised over the admissible set
\begin{equation*}
\mathcal{A}=\{\sigma\in L^\infty(\Omega):
\lambda\leq\sigma\leq\lambda^{-1}\mbox{ a.e. }\Omega
\},
\end{equation*}
for some $\lambda\in(0,1)$. The set
$\mathcal{A}$ is usually equipped with an $L^p(\Omega)$ norm $(1\leq p\leq \infty)$.
One may also employ data fitting other than the standard $L^2(\partial\Omega)$-norm. The most noteworthy one is the Kohn-Vogelius approach, which lifts the boundary data to the domain $\Omega$ and makes the fitting in $\Omega$ \cite{WexlerFryNeuman:1985,KohnMcKenney:1990,BGZ:03}; see also  \cite{Knowles:1998} for a
variant of the Kohn-Vogelius functional. In practice, the regularized formulations have to be properly 
discretized, commonly done by means of finite element methods \cite{gehre2014analysis,Rondi:2016,jin2017convergent,jin2019adaptive}, due to the spatially variable conductivity and irregular domain geometry. Newton-type methods have also been applied to EIT \cite{LechleiterRider:2006,LechleiterRieder:2008}.
Probabilistic formulations of these deterministic approaches are also possible \cite{KaipioKolehmainen:2000,gehre2014expectation,DunlopStuart:2016,Bohr:2023}, which can provide uncertainty estimates on the reconstruction.

The methods in the second group are of a more direct nature, aiming at extracting relevant information from the given data directly, without going through the expensive iterative process. Bruhl et al \cite{BruhlHanke:2000,BruhlVogeliusHanke:2003} developed the factorisation method for EIT, which provides a criterion for determining whether a point lies inside or outside the set of inclusions by carefully analysing the spectral properties of certain
operators. Thus, the inclusions can be reconstructed
directly by testing every point in the computational domain. The D-bar method of Siltanen, Mueller and Isaacson \cite{SiltanenMuellerIsaacson:2000,MuellerSiltanen:2020} is based on Nachman's uniqueness proof \cite{Nachman:1996} and utilises the complex geometric solutions and nonphysical scattering transform for direct image reconstruction. Chow, Ito and Zou \cite{chow2014direct} proposed the direct sampling method when there are only very few Cauchy data pairs. The method employs dipole potential as the probing function and constructs an indicator function for imaging the inclusions in EIT, and it is easy to implement and computationally cheap.
Other notable methods in the group include monotonicity method \cite{HarrachUllrich:2013}, enclosure method \cite{Ikehata:1998}, Calderón's method \cite{BikowskiMuller:2008,ShinMueller:2020}, and MUSIC \cite{AmmariLesselier:2005,AmmariGriesmaierHanke:2007,Lechleiter:2015} among others. Generally, direct methods are faster than those based on variational regularisation, but the reconstructions are often inferior in terms of resolution and can suffer from severe blurring. 

These represent the most common model-based inversion techniques for EIT reconstruction. Despite these important progress and developments, the quality of images
produced by EIT remains modest when compared with other imaging
modalities. In particular, at present, EIT reconstruction algorithms are still
unable to extract sufficiently useful information from data to
be an established routine procedure in many medical applications. Moreover, the iterative schemes are generally time-consuming, especially for 3D problems. One
possible way of improving the quality of information is to develop
an increased focus on identifying useful information and fully
exploiting a priori knowledge. This idea has been applied many times, and the recent advent of deep learning significantly expanded its horizon from hand-crafted regularisers to more complex and realistic learned schemes. Indeed, recently, deep learning-based approaches have been developed to address these challenges by drawing on knowledge encoded in the dataset or structural preference of the neural network architecture. 

We describe the sparsity approach and D-bar method next, and deep learning approaches in Section \ref{sec:DL}.

\subsection{Sparsity-based method} 

The sparsity concept is very useful for modelling conductivity distributions with ``simple'' descriptions away from the known background $\sigma_0$, e.g. when $\sigma$ consists of an uninteresting background plus
some small inclusions. Let $\delta\sigma^\dagger=\sigma^\dagger-\sigma_0$. A ``simple'' description
means that $\delta\sigma$ has a sparse representation with respect to a
certain basis/frame/dictionary $\{\psi_k\}$, i.e., there are only a few non-zero
expansion coefficients. The $\ell^1$ norm $\delta\sigma $ 
can promote the sparsity of  $\delta\sigma$  \cite{DDD:2004}
\begin{equation}\label{eqn:psi}
  \Psi(\sigma) = J(\sigma) + \alpha \|\delta\sigma\|_{\ell^1}, \quad\mbox{with}\quad\|\delta\sigma\|_{\ell^1}=\sum_{k}|\langle\delta\sigma,\psi_k\rangle|.
\end{equation} 
Under certain regularity conditions on $\{\psi_k\}$, the problem of minimising
$\Psi$ over the set $\mathcal{A}$ is well-posed \cite{jin2012analysis}.

Optimisation problems with the $\ell^1$ penalty have attracted intensive interest \cite{DDD:2004,BoneskyBrediesLorenzMaass:2007,BrediesLorenzMaass:2009,WrightNowak:2009}. The challenge lies in the non-smoothness of the $\ell^1$-penalty and high-degree nonlinearity of the
discrepancy $J(\sigma)$. The basic
algorithm for updating the increment $\delta \sigma_i$ and
$\sigma_i=\sigma_0+\delta \sigma_i$ by minimising $\Psi$ formally reads
\begin{equation*}
\delta \sigma_{i+1} = \mathcal{S}_{s\alpha}(\delta \sigma_{i}-s\Lambda_{\sigma_i,N}^{\prime\ast}(\Lambda_{\sigma_i,N}j-\phi^\delta)),
\end{equation*}
where $s>0$ is the step size, $\Lambda_{\sigma_i,N}^{\prime}$ denotes the G\^{a}teaux derivative of the NtD map $\Lambda_{\sigma_i,N}$ in $\sigma$, and $\mathcal{S}_\lambda(t)=\mbox{sign}(t)\max( |t|-\lambda,0)$ is the soft shrinkage operator.
However, a direct application of the algorithm does not
yield accurate results. We adopt the procedure in Algorithm \ref{alg:sparse}. The key tasks include computing the gradient $J'$ (Steps 4-5) and selecting the step size (Step 6).

\begin{algorithm}\small
\caption{Sparsity reconstruction for EIT.}
\label{alg:sparse}
\KwIn{ $\sigma_0$ and $\alpha$}
\KwResult{ an approximate minimiser $\delta\sigma$}
Set $\delta\sigma_0=0$\;
\For{i $\leftarrow$ 1, \ldots, I}{
    Compute $\sigma_{i}=\sigma_0+\delta\sigma_{i}$\;
    Compute the gradient $J'(\sigma_{i})$\;
    Compute the $H_0^1$-gradient $J'_s(\sigma_{i})$\;
    Determine the step size $s_{i}$\;
    Update inhomogeneity by $\delta\sigma_{i+1} = \delta\sigma_{i} - s_{i}J_s'(\sigma_{i})$\;
    Threshold $\delta\sigma_{i+1}$ by $\mathcal{S}_{s_i\alpha}(\delta\sigma_{i+1})$\;
    Check stopping criterion.
}
\end{algorithm}
\noindent\textbf{Gradient evaluation} Evaluating the gradient $J'(\sigma)=-\nabla u(\sigma)\cdot \nabla p(\sigma)$ involves solving an adjoint problem
\begin{equation*}
    -\nabla\cdot(\sigma\nabla p) = 0,\quad \mbox{in } \Omega, \quad \mbox{with}\quad \sigma\frac{\partial p}{\partial n} u(\sigma)-\phi^\delta \quad \mbox{on } \partial\Omega. 
\end{equation*}
Note that Indeed, $J'(\sigma)$ is defined via duality mapping
$J'(\sigma)[\lambda]=\langle
J'(\sigma),\lambda\rangle_{L_2(\Omega)}$,
and thus $J'(\sigma)\in (L^\infty(\Omega))'$ may be not smooth enough. Instead, we take
the $H_0^1(\Omega)$ metric for $\sigma$, by defining $J'_s(\sigma)$ via
$J'(\sigma)[\lambda]=\langle J_s'(\sigma),\lambda \rangle_{H_0^1(\Omega)}$.
Integration by parts yields
$-\Delta J_s'(\sigma)+J_s'(\sigma)=J'(\sigma)$ in $\Omega$ and 
$J_s'(\sigma) = 0$ on $\partial\Omega$. The assumption is that the inclusions are in the interior of $\Omega$. $J'_s$ is also known as
Sobolev gradient \cite{Neuberger:1997} and is a smoothed version of the
$L^2(\Omega)$-gradient. 
It metrises the set
$\mathcal{A}$ by the $H_0^1(\Omega)$-norm, thereby implicitly restricting the
admissible conductivity to a smoother subset. Numerically, evaluating the
gradient $J_s'(\sigma)$ involves solving a Poisson problem and can be carried out efficiently. Using  $J'_s$, we can locally approximate
$\Psi(\sigma)=\Psi(\sigma_0 + \delta \sigma)$ by
\begin{equation*}
\Psi(\sigma_0 + \delta \sigma) - \Psi(\sigma_0 + \delta \sigma_i) \sim
\langle\delta\sigma-\delta\sigma_i,J_s'(\sigma_i)\rangle_{H^1(\Omega)}
+\tfrac{1}{2s_i}\|\delta\sigma-\delta\sigma_i\|_{H^1(\Omega)}^2+\alpha\|\delta\sigma\|_{\ell^1},
\end{equation*}
which is equivalent to
\begin{equation}\label{eqn:proxy}
\tfrac{1}{2s_i}\|\delta\sigma-(\delta\sigma_i-s_iJ_s'(\sigma_i))
\|_{H^1(\Omega)}^2+\alpha\|\delta\sigma\|_{\ell^1}.
\end{equation}
Upon identifying $\delta\sigma$ with its expansion coefficients in $\{\psi_k\}$, the solution to problem \eqref{eqn:proxy} is given by
\begin{equation*}
\delta\sigma_{i+1}=\mathcal{S}_{s_i\alpha}(\delta\sigma_i-s_iJ_s'(\sigma_i)),
\end{equation*}
This step zeros out
small coefficients, thereby promoting the sparsity of $\delta\sigma$.

\noindent \textbf{Step size selection} Usually, gradient-type algorithms suffer from slow convergence, e.g., steepest descent methods.
One way to enhance its convergence is due to \cite{BarzilaiBorwein:1988}. The idea is to mimic the Hessian with $sI$ over the most recent steps so that
$sI(\delta\sigma_i-\delta\sigma_{i-1})\approx J_s'(\sigma_i)-J_s'(\sigma_{i-1})$
holds in a
least-squares sense, i.e.,
\begin{equation*}
s_i=\arg\min_s\|s(\delta\sigma_i-\delta\sigma_{i-1})-(
J_s'(\sigma_i)-J_s'(\sigma_{i-1}))\|_{H^1(\Omega)}^2. 
\end{equation*}
This gives rise to one popular Barzilai-Borwein rule
$s_i=\langle\delta\sigma_i-\delta\sigma_{i-1},J_s'
(\sigma_i)-J_s'(\sigma_{i-1})\rangle_{H^1(\Omega)}/\langle\delta\sigma_i
-\delta\sigma_{i-1},\delta\sigma_i-\delta\sigma_{i-1}\rangle_{H^1(\Omega)}$ \cite{BarzilaiBorwein:1988,DaiHagerZhang:2006}. In practice, following \cite{WrightNowak:2009}, we choose the step length $s$ to enforce a weak monotonicity
\begin{equation*}
\Psi(\sigma_0+\mathcal{S}_{s\alpha}(\delta\sigma_i-sJ_s'(\sigma_i)))\leq
\max_{i-M+1\leq k\leq i}\Psi(\sigma_k)-\tau\frac{s}{2}
\|\mathcal{S}_{s\alpha}(\delta\sigma_i-sJ'_s(\sigma_i))-\delta\sigma_i\|_{H^1(\Omega)}^2,
\end{equation*}
where $\tau$ is a small number, and $M\geq1$ is an integer. One may use the step size by the above
rule as the initial guess at each inner iteration and then decrease it geometrically by
a factor $q$ until the weak monotonicity is satisfied. The iteration is stopped when $s_i$ falls below a prespecified tolerance $s_\mathrm{stop}$ or when the maximum iteration number $I$  is reached.

The above description follows closely the work \cite{Jin_2012}, where the sparsity algorithm was first developed. There are alternative sparse reconstruction techniques, notably based on 
total variation \cite{RondiSantosa:2001,gehre2014analysis,BorsicLionheart:2010,ZhouHolder:2015}. For example, \cite{BorsicLionheart:2010} presented an experimental (in-vivo) evaluation of the total variation approach using a linearized model, and the resulting optimisation problem solved by the primal-dual interior point method; and the work \cite{ZhouHolder:2015} compared different optimisers.  Due to the non-smoothness of the total variation, one may relax the formulation with the Modica-Mortola function in the sense of Gamma convergence \cite{RondiSantosa:2001,jin2019adaptive}.

\subsection{The D-bar method} 
The D-bar method of Siltanen, Mueller and Isaacson \cite{SiltanenMuellerIsaacson:2000} is a direct reconstruction algorithm based on the uniqueness proof due to Nachman \cite{Nachman:1996}; see also Novikov \cite{Novikov:1988}. That is, a reconstruction is directly obtained from the DtN map $\Lambda_{\sigma,D}$, without going through an iterative process. Note that the DtN map $\Lambda_{\sigma,D}$ can be computed as the inverse of the measured NtD map $\Lambda_{\sigma,N}$ when full boundary data is available. Below we briefly overview the classic D-bar algorithm assuming $\sigma\in C^2(\Omega)$, with a positive lower bound (i.e., $\sigma \geq c>0$ in $\Omega$), and $\sigma\equiv 1$ in a neighbourhood of the boundary $\partial\Omega$. In this part, we consider an embedding of $\mathbb{R}^2$ in the complex plane, and hence we will identify planar points $x=(x_1,x_2)$ with the corresponding complex number $x_1+{\rm i}x_2$, and the product $kx$ denotes complex multiplication. For more detailed discussions, we refer interested readers to the survey \cite{MuellerSiltanen:2020}.

First, we transform the conductivity equation \eqref{eqn:eit} into a Schr\"odinger-type equation by substituting $\widetilde{u}=\sqrt{\sigma}u$ and setting $q=\Delta \sqrt{\sigma}/\sqrt{\sigma}$ and extending $\sigma\equiv 1$ outside $\Omega$. Then we obtain
\begin{equation}\label{eqn:Schrodinger}
    (-\Delta + q(x))\widetilde{u}(x) = 0,\quad  \mbox{in }\mathbb{R}^2.
\end{equation}
Next we introduce a class of special solutions of equation \eqref{eqn:Schrodinger} due to Faddeev \cite{Faddeev:1966}, the so-called complex geometrical optics (CGO) solutions $\psi(x,k)$, depending on a complex parameter $k\in\C \setminus \{0\}$ and $x\in\mathbb{R}^2$. These exponentially behaving functions are key to the reconstruction. Specifically, given $q\in L^p(\mathbb{R}^2), \, 1<p<2$, the CGO solutions $\psi(x,k)$ are defined as solutions to
\[
(-\Delta + q(x))\psi(\cdot,k) = 0,\quad \mbox{in }\mathbb{R}^2,
\]
satisfying the asymptotic condition $e^{-{\rm i}kx}\psi(x,k)-1 \in W^{1,\tilde{p}}(\R^2)$ with $2<\tilde{p}<\infty$. These solutions are unique for $k\in\C\setminus\{0\}$ as shown in \cite[Theorem 1.1]{Nachman:1996}. Then  D-bar algorithm recovers the conductivity $\sigma$ from the knowledge of the CGO solutions $\mu(x,k)=e^{-{\rm i}kx}\psi(x,k)$ at the limit $k\to 0$ \cite[Section 3]{Nachman:1996} 
\begin{equation*}
\lim_{k\to 0}\mu(x,k)    =\sqrt{\sigma},\quad x\in\Omega.
\end{equation*}
Numerically, one can substitute the limit by $k=0$ and evaluate $\mu(x,0)$.
The reconstruction of $\sigma$ relies on the use of an intermediate object called non-physical scattering transform $\mathbf{t}$, defined by 
\begin{equation*}\label{eqn:scatTrafo}
    \mathbf{t}(k) = \int_{\mathbb{R}^2}e_k(x)\mu(x,k)q(x){\rm d}x,
\end{equation*}
with $e_k(x):=\exp({\rm i}(kx+\bar k\bar x))$, where over-bar denotes complex conjugate. Since $\mu$ is asymptotically close to one, $\mathbf{t}(k)$ is similar to the Fourier transform of $q(x)$.  Meanwhile, we can obtain $\mu$ by solving the name-giving D-bar equation
\begin{equation}\label{eqn:Dbar}
\bar\partial_ k \mu(x,k)=\frac{1}{4\pi\bar{k}}\mathbf{t}(k)e_{-k}(x)\overline{\mu(x,k)},\quad k\neq 0,
\end{equation}
where $\bar\partial_k =\frac12 (\frac{\partial}{\partial k_1}+{\rm i}\frac{\partial}{\partial k_2})$ is known as the D-bar operator.
To solve the above equation, scattering transform $\mathbf{t}(k)$ is required, which we can not measure directly from the experiment, but $\mathbf{t}(k)$ can be represented using the DtN map. Indeed, using Alessandrini's identity \cite{Alessandrini}, we get the boundary integral
\[
\mathbf{t}(k)=\int_{\partial\Omega} e^{{\rm i}\bar{k}\bar{x}}(\Lambda_{\sigma,D}-\Lambda_{1,D}) \psi(x,k) {\rm d}s.
\]
Note that $\Lambda_{1,D}$ can be analytically computed, and only $\Lambda_{\sigma,D}$ needs to be obtained from the measurements. Here, we will employ a Born approximation using $\psi\approx e^{{\rm i}kx}$, leading to the linearised approximation
\begin{equation}\label{eqn:bornApprox}
\mathbf{t}^{\exp}(k)\approx\int_{\partial\Omega} e^{{\rm i}\bar{k}\bar{x}}(\Lambda_{\sigma,D}-\Lambda_{1,D}) e^{{\rm i}kx} {\rm d}s.
\end{equation}
This linearised D-bar algorithm can be efficiently implemented. First, one computes the $\mathbf{t}^{\exp}(k)$ from the measured DtN map $\Lambda_{\sigma,D}$, and then one solves the D-bar equation \eqref{eqn:Dbar}. Note that the solutions of \eqref{eqn:Dbar} are independent for each $x\in\Omega$ and one can efficiently parallelise over $x$. This leads to real-time implementations and is especially relevant for time-critical applications, e.g., monitoring purposes.  The fully nonlinear D-bar algorithm would require first computing $\psi$ by solving a boundary integral equation and then computing the scattering transform $\mathbf{t}(k)$.

The above algorithm assumes infinite precision and noise-free data. When the data is noise corrupted with finite measurements, the measured DtN map $\Lambda_{\sigma,D}$ is not accurate, and then the computation of $\mathbf{t}(k)$ becomes exponentially unstable for $|k|>R$. Thus, for practical data, we need to restrict the computations to a certain frequency range so as to stably compute $\mathbf{t}(k)$. Below we choose $R=5$ for noise-free data and $R=4.5,\, R=4$ for 1\% and 5\% noisy measurements, respectively. This strategy of reducing the cut-off radius for noisy measurements is shown to be a regularisation strategy \cite{KnudsenLassasSiltanen:2009}. The final algorithm can be summarised as outlined below in Algorithm \ref{alg:Dbar}.

\begin{algorithm}\small
\caption{D-bar algorithm using $\mathbf{t}^{\exp}$}
\label{alg:Dbar}
\KwIn{ $\Lambda_{\sigma,D}$ and $R$}
\KwResult{Regularised reconstruction of $\sigma$}
Compute analytic $\Lambda_{1,D}$\;
Evaluate $\mathbf{t}^{\exp}(k)$ for $|k|<R$ by \eqref{eqn:bornApprox}\;
Solve the D-bar equation \eqref{eqn:Dbar}\;
Obtain $\sigma(x) = \mu(x,0)^2$ for $x\in\Omega$\;
\end{algorithm}
Besides the D-bar method, there are other analytic and direct reconstruction methods available, e.g., enclosure method \cite{Ikehata:1998}, monotonicity method \cite{HarrachUllrich:2013}, direct sampling method \cite{chow2014direct}, and Calder\'on's method \cite{BikowskiMuller:2008,ShinMueller:2020}. The common advantage of these approaches is their computational efficiency, but unfortunately, also the directly inherited exponential instability to noise. While there are strategies to deal with noise, e.g., reducing the cut-off radius, the reconstruction quality does suffer: the reconstructions tend to be overall smooth. Additionally, there may be theoretical limitations to the reconstructions that can be obtained. For example, for the classic D-bar algorithm, it is  $C^2$ conductivities, and for the enclosure methods, we can only find the convex hull of all inclusions.
Thus, it is very interesting to discuss how deep learning can help overcome these limitations. 


\section{Deep learning-based methods}\label{sec:DL}

The integration of deep learning techniques has significantly advanced EIT reconstruction. It has successfully addressed several challenges posed by the non-linearity and severe ill-posedness of the inverse problem, leading to improved quality and reconstruction accuracy. Researchers have achieved breakthroughs in noise reduction, edge retention, and spatial resolution, making EIT a more viable imaging modality in medical and industrial applications. This success is mainly attributed to the extraordinary approximation ability of DNNs and the use of a large amount of paired training datasets. 

First, much effort has been put into designing DNNs architectures for directly learning the maps from the measured voltages $U$ to conductivity distributions $\sigma$, i.e., training a DNN $\mathcal{G}_\theta$ such that $\sigma\approx \mathcal{G}_\theta(U)$. 
Li et al. \cite{li2017image} proposed a four-layer DNN framework constituted of a stacked autoencoder and a logistic regression layer for EIT problems. Tan et al. \cite{tan2018image} designed the network based on LeNet convolutional layers and refined it using pooling layers and dropout layers. Chen et al. \cite{9128764} introduce a novel DNN using a fully connected layer to transform the measurement data to the image domain before a U-Net architecture, and  \cite{yang2023eit} a DenseNet with multiscale convolution. Fan and Ying \cite{fan2020solving} proposed DNNs with compact architectures for the forward and inverse problems in 2D and 3D, exploiting the low-rank property of the EIT problem. Huang et al. \cite{huang2019improved} first reconstruct an initial guess using RBF networks, which is then fed into a U-Net for further refinement. \cite{seo_harrach_2019}, uses a variational autoencoder to obtain a low-dimensional representation of images, which is then mapped to a low dimension of the measured voltages as well. We refer to \cite{wu2021shape,li2020one,ren2021rcrc,chen2021structure} for more direct learning methods. 

Second, combining traditional analytic-based methods and neural networks is also a popular idea. Abstractly, one employs an analytic operator $\mathcal{R}$ and a neural network $\mathcal{G}_\theta$ such that $\sigma\approx\mathcal{G}_\theta(\mathcal{R}(U))$. One example is the Deep D-bar method \cite{hamilton2018deep}. It first generates EIT images by the D-bar method, then employs the U-Net network to refine the initial images further. Along this line, one can design the input of the DNN  from Calder\'{o}n's method \cite{cen2023electrical,SunZhongWang:2023}, domain-current method \cite{wei2019dominant}, one-step Gauss-Newton algorithm  \cite{martin2017post} and conjugate gradient algorithm \cite{zhang2022v}. Inspired by the mathematical relationship between the Cauchy difference index functions in the direct sampling method, Guo and Jiang \cite{guo2021construct} proposed the DDSM proposed in \cite{guo2021construct} employs the Cauchy difference functions as the DNN input. Yet another popular class of deep learning-based methods that combines model-based approaches with learned components is based on the idea of unrolling, which replaces components of a classical iterative reconstructive method with a neural network learned from paired training data (see \cite{MongaEldar:2021} for an overview). Chen et al. \cite{ZhouYang:2022} proposed a multiple measurement vector (MMV) model-based learning algorithm (called MMV-Net) for recovering the frequency-dependent conductivity in multi-frequency electrical impedance tomography (mfEIT). It unfolds the update steps of the alternating direction method of multipliers for the MMV problem. The authors validated the approach on the Edinburgh mfEIT Dataset and a series of comprehensive experiments. See also \cite{ZhouYang:2022b} for a mask-guided spatial–temporal graph neural network (M-STGNN) to reconstruct mfEIT images in cell culture imaging.
Unrolling approaches based on the Gauss-Newton have also been proposed, where an iterative updating network is learned for the explicitly computed Gauss-Newton updates \cite{Herzberg:2021} or a proximal type operator \cite{colibazzi2023deep}. Likewise, a quasi-Newton method has been proposed by learning an updated singular value decomposition \cite{smyl2021efficient}. One should further mention an excellent study on how to apply deep learning concepts for the particular case of EIT-lung data \cite{seo_harrach_2019}, which sets the standards in terms of integrating mathematical as well as clinical expertise into the learned reconstruction process.

Reconstruction methods in these two groups are supervised in nature and rely heavily on high-quality training data. Even though there are a few public EIT datasets, they are insufficient to train DNNs (often with many parameters). In practice, the DNN is learned on synthetic data, simulated with phantoms via, e.g., FEM. The main advantage is that once the neural network is trained, at the inference stage, the process requires only feeding through the trained neural network and thus can be done very efficiently. Generally, these approaches perform well when the test data is close to the distribution of the training data. Still, their performance may degrade significantly when the test data deviates from the setting of the training data \cite{Antun:2020}. This lack of robustness with respect to the out-of-distribution test data represents one outstanding challenge with all the above approaches.

Third, several unsupervised learning methods have been proposed for EIT reconstruction. Bar et al. \cite{bar2021strong} employ DNNs to approximate voltage functions $\{u_j\}_{j=1}^J$ and conductivity $\sigma$ and then train them together to satisfy the strong PDE conditions and the boundary conditions, following the physics-informed neural networks (PINNs) \cite{raissi2019physics}. Furthermore, data-driven energy-based models are imparted onto the approach to improve the convergence rate and robustness for EIT reconstruction  \cite{pokkunuru2022improved}. Bao et al. \cite{bao2020numerical} exploited the weak formulation of the EIT problem, using DNNs to parameterise the solutions and test functions and adopting a minimax formulation to alternatively update the DNN parameters (to find an approximate solution of the EIT problem). Liu et al. \cite{liu2023deepeit} applied the deep image prior (DIP) \cite{ulyanov2018deep}, a novel DNN-based approach to regularise inverse problems, to EIT, and optimised the conductivity function by back-propagation and the finite element solver. Generally, the methods in this group are robust with respect to the distributional shift of the test data. However, each new test data requires fresh training, and hence, they tend to be computationally more expensive.

In addition, several neural operators, e.g., \cite{lu2021learning,li2020fourier,tripura2022wavelet}, have been designed to approximate mostly forward operators. The recent survey \cite{Tanyu_2023} discusses various extensions of these neural operators for solving inverse problems by reversed input-output and studied Tikhonov regularisation with a trained forward model.

\subsection{Deep D-bar} 
In practice, reconstructions obtained with the D-bar method suffer from a smoothing effect due to truncation in the scattering transform, which is necessary for finite and noisy data but leaves out all high-frequency information in the data. Thus, we cannot reconstruct sharp edges, and subsequent processing is beneficial. An early approach to overcome the smoothing is to use a nonlinear diffusion process to sharpen edges \cite{HamiltonHauptmannSiltanen:2014}. In recent years, deep learning has been highly successful for post-processing insufficient noise or artefact-corrupted reconstruction \cite{JinUnser:2017}. 

In the context of the deep D-bar method, we are given an initial analytic reconstruction operator $\mathcal{R}_{\text{d-bar}}$ that maps the measurements (i.e., the DtN map $\Lambda_{\sigma,D}$ for EIT) to an initial image, which suffers from various artefacts, primarily over-smoothing. Then a U-Net $\mathcal{G}_\theta$ \cite{ronneberger2015u} is trained to improve the reconstruction quality of the initial reconstructions, and we refer to the original publication \cite{hamilton2018deep} for details on the architecture. Thus, we could write this process as $\sigma \approx \mathcal{G}_\theta(\mathcal{R}_{\text{d-bar}}(\Lambda_{\sigma,D}))$, where the network $\mathcal{G}_\theta$ is trained by minimising the $\ell^2$-loss of D-bar reconstructions to ground-truth images. Specifically, given a collection of $N$ paired training data $\{(\sigma_i^\dag,\Lambda_{\sigma_i^\dag,D}^\delta)\}_{i=1}^N$ (i.e., ground-truth conductivity $\sigma_i^\dag$ and the corresponding noisy measurement data $\Lambda_{\sigma_i^\dag,D}^\delta$), we train a DNN $\mathcal{G}_\theta$ by minimising the following empirical loss
\begin{equation*}
    \mathcal{L}(\theta) = \frac{1}{N}\sum_{i=1}^N\|\sigma_i^\dag-\mathcal{G}_\theta( \mathcal{R}_{\text{d-bar}}(\Lambda_{\sigma_i^\dag,D}^\delta))\|_{L^2(\Omega)}^2,
\end{equation*}
This can be viewed as a specialised denoising scheme to remove the artefacts in the initial reconstruction
$\mathcal{R}_\text{d-bar}(\Lambda_{\sigma_i^\dag,D}^\delta)$ by the D-bar reconstructor $\mathcal{R}_\text{d-bar}$. The loss $\mathcal{L}(\theta)$ is then minimised with respect to the DNN parameters $\theta$, typically by the Adam algorithm \cite{KingmaBa:2015}, a very popular variant of stochastic gradient descent. Once a minimiser $\theta^*$ of the loss $\mathcal{L}(\theta)$ is found, given a new test measurement $\Lambda_{\sigma,D}^\delta$, we can obtain the reconstruction $\mathcal{G}_{\theta^*}(\mathcal{R}_\text{d-bar}(\Lambda_{\sigma,D}^\delta))$. Thus at the testing stage, the method requires only additional feeding of the initial reconstruction $\mathcal{R}_\text{d-bar}(\Lambda_{\sigma,D}^\delta)$ through the network $\mathcal{G}_{\theta^*}$, which is computationally very efficient. This presents one distinct advantage of a supervisedly learned map.

Several extensions have been proposed. Firstly, the need to model boundary shapes in the training data can be eliminated by using the Beltrami approach \cite{AstalaPaivrinta:2006} instead of the classic D-bar method. This allows for domain-independent training \cite{Hamilton:2019Kolehmainen:2019}. A similar motivation is given by replacing the classic U-net that operates on rectangular pixel domains with a graph convolutional version; this way learned filters are domain and shape-independent \cite{Herzberg:2021,herzberg2023domain}. Similarly, the reconstruction from Calder\'on's method \cite{BikowskiMuller:2008,ShinMueller:2020} can be post-processed using U-net, leading to the deep Calder\'{o}n's method \cite{cen2023electrical}. Distinctly, the deep Calder\'on's method is capable of directly recovering complex valued conductivity distributions. Finally, even the enclosure method can be improved by predicting the convex hull from values of the involved indicator function \cite{siltanen2020electrical}.


\subsection{Deep direct sampling method}
The deep sampling method (DDSM) \cite{guo2021construct} is based on the direct sampling method (DSM) due to Chow, Ito and Zou \cite{chow2014direct}. Using only one single Cauchy data pair on the boundary $\partial\Omega$, The DSM constructs a family of probing functions $\{\eta_{x,d_x}\}_{x\in \Omega,d_x\in \mathbb{R}^n}\subset H^{2\gamma}(\partial\Omega)$ such that the index function defined by
\begin{equation}
	\mathcal{I}(x,d_x):=\frac{\langle \eta_{x,d_x},u-u_{\sigma_0} \rangle_{\gamma,\partial\Omega}}{\Vert u-u_{\sigma_0}\Vert_{L^2(\partial\Omega)}|\eta_{x,d_x}|_Y},\quad x\in \Omega,
\end{equation}
takes large values for points near the inclusions and relatively small values for points far away from the inclusions, where $|\cdot|_Y$ denotes the $H^{2\gamma}(\partial\Omega)$ seminorm in  and the duality product $\langle f,g \rangle_{\gamma,\partial\Omega}$ is defined by 
\begin{equation}
\langle f,g\rangle_{\gamma,\partial\Omega}=\int_{\partial\Omega}(-\Delta_{\partial\Omega})^{\gamma}fg{\rm d}s=\langle (-\Delta_{\partial\Omega})^{\gamma}f, g\rangle_{L^2(\partial\Omega)},	
\end{equation}
where $-\Delta_{\partial\Omega}$ denotes the Laplace-Beltrami operator, and $(-\Delta_{\partial\Omega})^\gamma$ its fractional power via spectral calculus. Let the Cauchy difference function $\varphi$ be defined by
\begin{equation}
	-\Delta \varphi=0 \quad\text{in}\quad \Omega, \quad \frac{\partial\varphi}{\partial n}=(-\Delta_{\partial\Omega})^\gamma(u_\sigma-u_{\sigma_0}) \quad \text{on}\quad \partial\Omega, \quad\int_{\partial\Omega} \varphi {\rm d}s=0.
\end{equation}
Then the index function $\mathcal{I}(x,d_x)$ can be equivalently rewritten as
\begin{equation}
	\mathcal{I}(x,d_x):=\frac{d_x\cdot\nabla \varphi(x)}{\Vert u_\sigma-u_{\sigma_0}\Vert_{L^2(\partial\Omega)}|\eta_{x,d_x}|_Y},\quad x\in \Omega,
\end{equation}

Motivated by the relation between the index function $\mathcal{I}(x,d_x)$ and the Cauchy difference function $\varphi$ and to fully make use of multiple pairs of measured Cauchy data, Guo and Jiang \cite{guo2021construct} proposed the DDSM, employing DNNs to learn the relationship between the Cauchy difference functions $\varphi$ and the true inclusion distribution. That is, DSSM construct and train a DNN $\mathcal{G}_\theta$ such that 
\begin{equation}
	\sigma\approx\mathcal{G}_\theta(\varphi_1,\varphi_2,...,\varphi_N),
\end{equation}   
where $\{\varphi_i\}_{i=1}^{N}$ correspond to $N$ pairs of Cauchy data  $\{g_\ell,\Lambda_{\sigma,N} g_\ell\}_{\ell=1}^N$.
Guo and Jiang \cite{guo2021construct} employed a CNN-based U-Net network for DDSM, and later  \cite{guo2022transformer} designed a U-integral transformer architecture (including comparison with state-of-the-art DNN architectures, e.g., Fourier neural operator, and U-Net). In our numerical experiments, we choose the U-Net as the network architecture for DDSM as we observe that U-Net can achieve better results than the U-integral transformer for resolution $64 \times 64$. For higher resolution cases, the U-integral transformer seems to be a better choice due to its more robust ability to capture long-distance information. 
The following result \cite[Theorem 4.1]{guo2021construct} provides some mathematical foundation of DDSM.
\begin{thm}
Let $\{g_\ell\}_{\ell=1}^\infty$ be a fixed orthonormal basis of $H^{-1/2}(\partial\Omega)$. Given
an arbitrary $\sigma$ such that $\sigma>\sigma_0$ or $\sigma<\sigma_0$, let $\{g_\ell,\Lambda_{\sigma,N} g_\ell\}_{\ell=1}^\infty$ be the Cauchy data pairs and let $\{\varphi_\ell\}_{\ell=1}^\infty$ be the corresponding Cauchy difference functions with $\gamma=\sigma_0$. Then the inclusion distribution $\sigma$ can be purely determined from $\{\varphi_\ell\}_{\ell=1}^\infty.$
\end{thm}

The idea of DDSM was extended to diffusive optical tomography in \cite{guo2023learn}. Ning et al. \cite{ning2023direct} employ the index functions obtained from the DSM as the input of the DNN for solving inverse obstacle scattering problems. 


\subsection{CNN based on LeNet}

\begin{figure}[hbt!]
	\centering
	\includegraphics[width=0.7\textwidth]{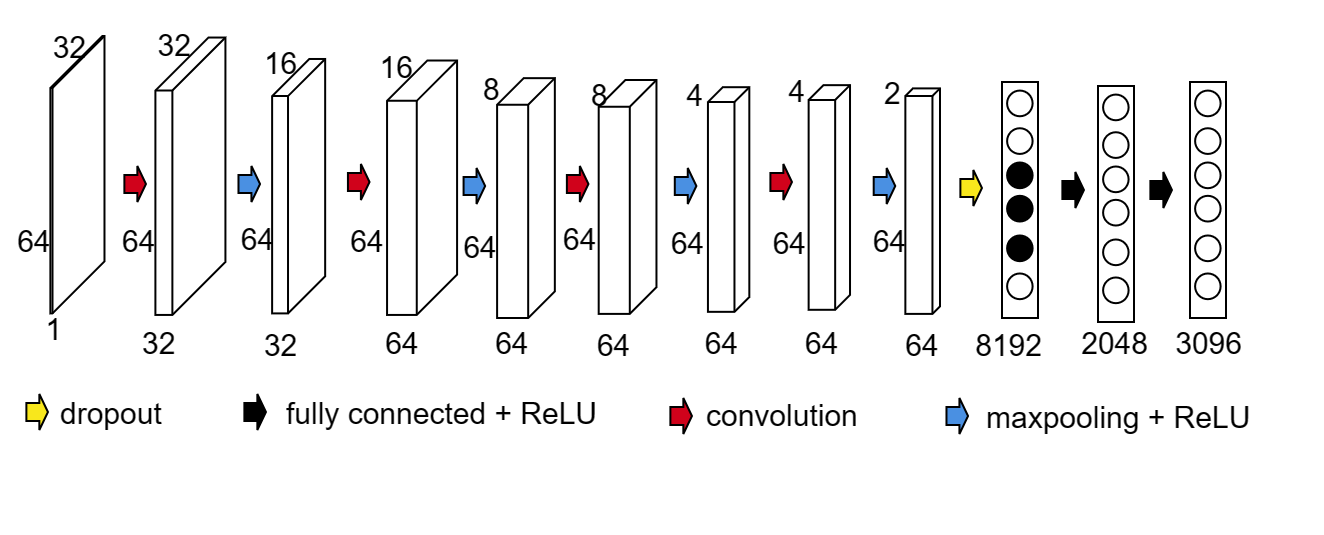}
	\caption{The architecture of CNN-based on LeNet.}
	\label{fig:LeNet}
\end{figure}

Li et al. \cite{li2017image} proposed using CNN to directly learn the map from the measured data and the conductivity distribution. The employed network architecture is based on LeNet and refined by applying dropout layer and
moving average. The CNN architecture used in the numerical experiments below is shown in Fig. \ref{fig:LeNet}. Since the number of injected currents and the discretisation size differ from that in \cite{li2017image}, we modify the input size, network depth, kernel size, etc. The input size is 32 × 64. The kernel size is 5 × 5 with zero-padding max pooling rather than average pooling is adopted to gain better performance. The sigmoid activation function used in LeNet causes a serious saturation phenomenon, which can lead to vanishing gradients. So, ReLU is chosen as the activation function below. A dropout layer is added to improve the generalisation ability of this model. One-half of the neurons before the first fully connected layers are randomly discarded from the network during the training process. It can reduce the complex co-adaptation among neurons so that the network can learn more robust features. In addition, a dropout layer has been proven to be very effective in training large datasets.

\subsection{FC-UNet} 

Chen et al. \cite{9128764} proposed a novel deep learning architecture by adding a fully connected layer before the U-Net structure. The input of the network is given by the difference voltage $u_\sigma^\delta-u_{\sigma_0}$. Inspired by a linearized approximation of the EIT problem for a small perturbation of conductivity
distribution $\sigma-\sigma_0$:
\begin{equation}
    u_\sigma^\delta-u_{\sigma_0}\approx \mathbf{J}(\sigma-\sigma_0),
\end{equation}
where $\mathbf{J}$ donates the sensitivity matrix, the method first generates an initial guess of the conductivity distribution $\sigma$ from the linear fully connected(FC) layer followed by a ReLU layer and then feeds it to a denoising U-Net model to learn the nonlinear relationship further. Thus we could write this process as $\sigma\approx \mathcal{G}_\theta(u_\sigma^\delta-u_{\sigma_0})=\mathcal{G}_{\theta_2}(\mathcal{G}_{\theta_1}(u_\sigma^\delta-u_{\sigma_0}))$ with $\mathcal{G}_{\theta_1}=\text{FC+ReLU}$ and $\mathcal{G}_{\theta_2}=\text{U-Net}$. The authors also proposed an initialisation strategy to further help obtain the initial guess, i.e., the weights $\theta_1$ of the fully connected layer are initialised with the least-squares solution using training data. The weights $\theta_2$ for the U-Net are initialised randomly as usual. Then, all weights $\theta=\theta_1\cup\theta_2$ are updated during the training process. According to the numerical results shown in \cite{9128764}, this special weight initialization
strategy can reduce the training time
and improve the reconstruction quality. With a trained network, different from the deep D-bar and DDSM methods,  the methods FC-UNet and CNN based on LeNet only involve a forward pass of the trained network for each testing example. 

\begin{figure}[ht!]
	\centering	\includegraphics[width=0.8\textwidth]{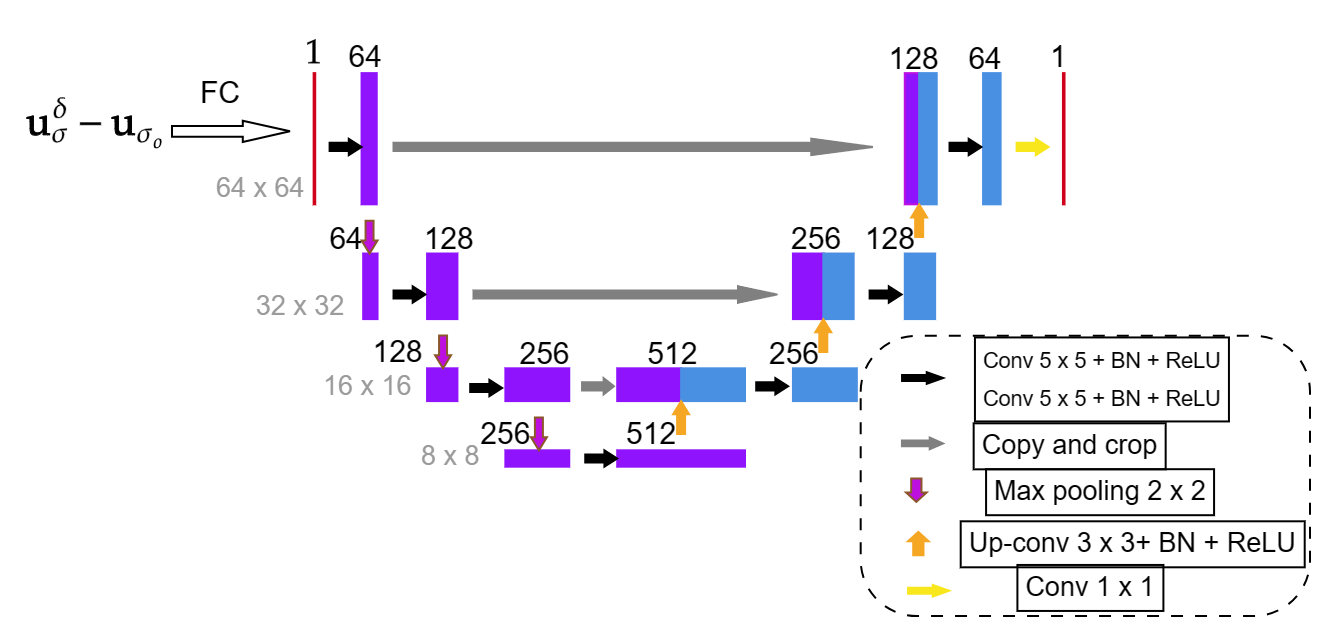}
	\caption{The architecture of FC-UNet.}
	\label{fig:FCUNET}
\end{figure}

Based on our numerical experience, dropping the ReLU layer following the fully connected layer can provide better reconstruction results, at least for the examples in section \ref{sec:experiment}. Thus, for the numerical experiments, we employ the FC-UNet network as shown in Fig. \ref{fig:FCUNET}, in which only a linear fully connected layer is employed before the U-Net.

In addition, by employing the FC-UNet to extract structure distribution
and a standard CNN to extract conductivity values, a structure-aware dual-branch network was designed in \cite{chen2021structure} to solve EIT problems.  

\newpage
\section{Numerical experiments and results}\label{sec:experiment}
The core of this work is the extensive numerical experiments. Now, we describe how to generate the dataset used in the experiments, highlighting its peculiarity and relevance in real-world scenarios, and also the performance metrics used for comparing different methods. Last, we present and discuss the experimental results.
\subsection{Dataset  generation and characteristics}
Generating simulated data consists of three main parts, which we describe below. The codes for data generation are available at 
\url{https://github.com/dericknganyu/EIT_dataset_generation}.

In the 2D setting, we generate $N$ circular phantoms $\{P_i\_{i = 1}^N$, all restricted to the unit circle centred at the origin, i.e., $\Omega=\{ (x, y):x^2 + y^2 \leq 1\}$ in the Cartesian coordinates or $\{(r, \theta):r \leq 1, \theta \in [0, 2\pi]\}$ in polar coordinates. The phantoms are generated randomly. Firstly, we decide on the maximum number  $M \in \mathbb{N}$ of inclusions. Each phantom would then contain $n$ inclusions, where $n \in \mathcal{U}\{1,\ldots, M\}$, the uniform distribution over the set $[1,\ldots, M]$. To mimic realistic scenarios in medical imaging, the inclusions are elliptical and are sampled such that when $n>1$, the inclusions do not overlap. Since the inclusions are elliptical, each inclusion, $(E_j)_{j = 1}^n$ is characterised by a centre $C_j = (h_j, k_j)$, an angle of rotation $\alpha_j$, a major and minor axis $a_j$ and $b_j$ respectively. The parametric equation of an ellipse $E_j$ is thus given by
\begin{equation}
    E_j = \left\{ (x, y): \begin{pmatrix}
                    x\\
                    y
                \end{pmatrix} = 
                \begin{pmatrix}
                    h_j + a_j\cos{\theta}\cos{\alpha_j} - b_j\sin{\theta}\sin{\alpha_j}\\
                    k_j + a_j\cos{\theta}\sin{\alpha_j} + b_j\sin{\theta}\cos{\alpha_j}
                \end{pmatrix}, \theta  \in [0, 2\pi]  \right\}.
                \label{eqn:param-ellipse}
\end{equation} 
To mimic realistic scenarios in medical imaging, the inclusions are sampled to avoid contact with the boundary  $\partial \Omega$ of the domain $\Omega$. For an inclusion $E_j$, we have $x^2 + y^2 < 0.9$ for any $(x, y) \in E_j$. In this way, all phantoms have inclusions contained within $\Omega$. We illustrate this in Algorithm \ref{algo:gen_phan}.

\begin{figure}[!ht]
	\centering
	\begin{subfigure}[b]{0.3\textwidth}
		\centering
		\includegraphics[width=0.9\textwidth]{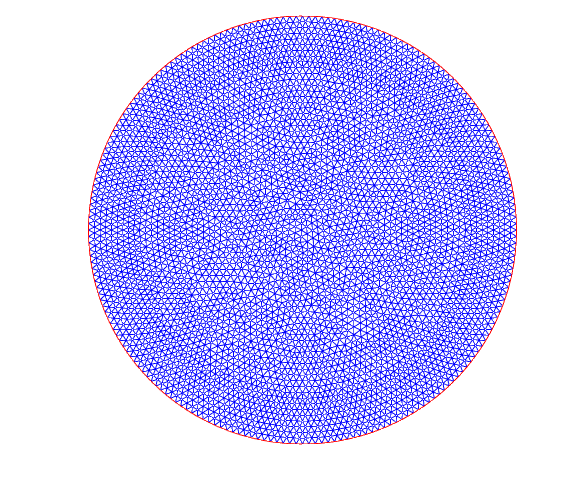}
		\caption{Forward solver FEM mesh.}
		\label{fig:mesh}
	\end{subfigure}
	\hfill
	\begin{subfigure}[b]{0.3\textwidth}
		\centering
		\includegraphics[width=0.9\textwidth]{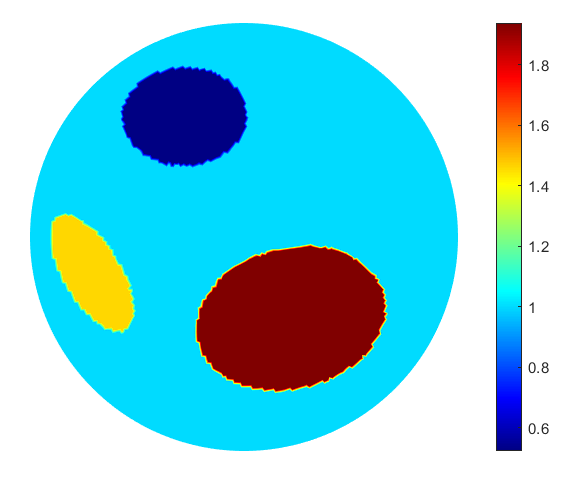}
		\caption{Constant $\sigma^i_j\big|^3_{j = 1}$ inclusions.}
		\label{fig:sigma_constant}
	\end{subfigure}
	\hfill
	\begin{subfigure}[b]{0.3\textwidth}
		\centering
		\includegraphics[width=0.9\textwidth]{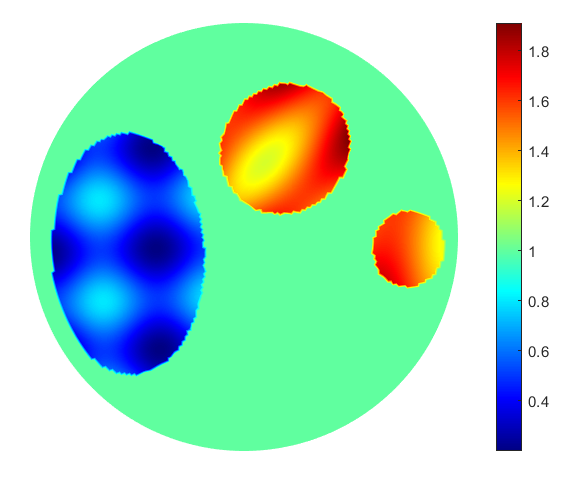}
		\caption{Textured $\sigma^i_j\big|^3_{j = 1}$ inclusions.}
		\label{fig:sigma_texture}
	\end{subfigure}
	\caption{Illustrating of Phantom characteristics used in simulated data.}
	\label{fig:sigmaa}
\end{figure}

Each phantom $P_i, i \in \{1, 2, \ldots, N\},$ has $(E_j)_{j = 1}^n$ inclusions, with $n \in \mathcal{U}\{1,\ldots, M\}$. For each $E_j$, we assign a conductivity $\sigma^i_j \in \Sigma_j := \mathcal{U}(0.2, 0.8) \cup \mathcal{U}(1.2, 2.0)$. The background conductivity is set to $1$. In this way, given a point $(x, y) \in P_i$ in the domain/phantom, the conductivity $\sigma_i(x,y)$ at that point is therefore given by
\begin{equation}
    \sigma_i (x, y) =
        \begin{cases}
          \sigma^i_j \in \Sigma_j , & \text{if}\ (x, y) \in E_j,~ j = 1, \ldots, n \\
          1, & \text{otherwise.}
        \end{cases}
    \label{eqn:sigma_constant}
\end{equation}
Fig. \ref{fig:sigma_constant} shows an example of a phantom generated in this way.

Next, for any simulated $\sigma$, we solve the forward problem \eqref{eqn:eit} using the Galerkin finite element method (FEM) \cite{LechleiterRider:2006,gehre2014analysis}, for the injected currents $g_1$ and $g_2$ in \eqref{eqn:g} around the boundary $\partial \Omega$. The points $(x, y) \in P_i$ are thus nodes in the finite element mesh shown in Fig. \ref{fig:mesh} \begin{equation}
	\begin{aligned}
        g_1 &= \pi^{-1/2} \sin (n \theta)\quad \mbox{and}\quad
        g_2 &= \pi^{-1/2} \cos (n \theta), \quad n=1,2, \ldots, 16
	\end{aligned}
    \label{eqn:g}
\end{equation}
We use the MATLAB PDE toolbox in the numerical experiment to solve the forward problem.
\begin{algorithm}\small
\caption{Procedure for generating phantoms}
\label{algo:gen_phan}
\KwIn{\\
	    \qquad$\bullet$ nodes $(x_\ell, y_\ell), \ell \in \{1, 2, \ldots, L\}$ from FEM mesh \\
	    \qquad$\bullet$ $N \in \mathbb{N}$, number of phantoms\\
	    \qquad$\bullet$ $M \in \mathbb{N}$, maximum number of inclusions
     }
\KwResult{\\
	    \qquad$\bullet$ Phantoms $P_i$, with conductivity $\sigma_i, i \in \{1, 2, \ldots, n\}$\\
     }
select $n \in \mathcal{U}\{1, M\}$\;
\For{i $\leftarrow$ 1, \ldots, N}{
    \For{j $\leftarrow$ 1, \ldots, n}{
        \tcc{Sample inclusions and conductivity}
        Sample $E_j$, non-overlapping ellipses  based on \eqref{eqn:param-ellipse}, within the circle of radius $0.9$\;
        Sample $\sigma^i_j \in \mathcal{U}(0.2, 0.8) \cup \mathcal{U}(1.2, 2.0)$\;
    }
    \For{$\ell \leftarrow$ 1, \ldots, L}{
        \tcc{Evaluate conductivity on mesh nodes}
       Evaluate $\sigma_i(x_\ell, y_\ell)$ based on \eqref{eqn:sigma_constant} \;
    }
}
\end{algorithm}
 
In real-life situations, the conductivities of the inclusions are rarely constant. Indeed, usually, there are textures on internal organs in medical applications. Motivated by this, we take a step further in generating phantoms, with inclusions having variable conductivities. This introduces a novel challenge to the EIT problem, and we seek to study its impact on different reconstruction algorithms. The procedure to generate simulated data remains unchanged. However, $\sigma^i_j$ in equation \eqref{eqn:sigma_constant} becomes
\begin{equation*}
    \sigma^i_j = s \circ f \circ R_{\alpha_j,C_j}, 
\end{equation*}
where $f : \mathbb{R}^2 \ni (x, y) \mapsto \tfrac{1}{2} \left(\sin{k_x x} + \sin{k_y y}\right) \in [-1, 1]$, $R_{\alpha_j,C_j}$ is the rotation of centre $C_j = (h_j, k_j)$ and angle $\alpha_j$, with respect to the centre and angle of the ellipse $E_j$ respectively; and $s$ applies a scaling so that the resulting $\sigma^i_j$ is either within the range $[0.2, 0.8]$ or $[1.2, 2.0]$. Fig. \ref{fig:sigma_texture} shows an example phantom. 

We also study the performance of the methods in noisy scenarios, i.e. reconstructing the conductivity from noisy measurements. The resulting solution to the forward problem $u$, on the boundary $\partial\Omega$, is then perturbed with normally distributed random noise of different levels $\delta$:  
\begin{equation*}
    u^\delta(x) = u(x) + \delta \cdot |u(x)| \cdot \xi(x) ,  \quad x \in \partial\Omega,
\end{equation*}
where $\xi(x)$ follows the standard normal distribution $\mathcal{N}(0,1)$. 

For the deep learning methods, we employ 20,000 training data and 100 validation data without noises added. Then we compare the results for 100 testing data with different noise levels.

We employ several performance metrics commonly used in the literature to compare different reconstruction methods comprehensively. Table \ref{tab:perf_metric} outlines these metrics with their mathematical expressions and specifications. In Table \ref{tab:perf_metric}, $\boldsymbol{\sigma}$ denotes the ground truth with mean $\mu_{\boldsymbol{\sigma}}$ and variance $s^2_{\boldsymbol{\sigma}}$, while $\boldsymbol{\hat{\sigma}}$ the predicted conductivity  with mean $ \mu_{\boldsymbol{\hat{\sigma}}}$ and variance $ s^2_{\boldsymbol{\hat{\sigma}}}$. $\hat{\sigma}_i$ is the $i$-th element of $ \boldsymbol{\hat{\sigma}}$ while  $\sigma_i$ is the $i$-th element of $ \boldsymbol{\sigma}$. ${N}$ is the total number of pixels, so that $\boldsymbol{\sigma}= (\sigma_i)^N_{i=1}$ and $\boldsymbol{\hat\sigma}= (\hat 
 \sigma_i)^N_{i=1}$.

\begin{table}[ht!]\small        
        \makegapedcells
	\centering
        \begin{tabular}{@{} >{\centering\arraybackslash}m{0.15\linewidth} >{\centering\arraybackslash}m{0.4\linewidth} m{0.35\linewidth} @{}}
        \toprule
        Error Metric & Mathematical Expression & Highlights  \\\midrule
        Relative Image Error (RIE) & $\dfrac{\| \hat{\boldsymbol{\sigma}}- \boldsymbol{\sigma}\|}{\| \boldsymbol{\sigma}\|} = \dfrac{\sum\limits_{i=1}^N\left|\hat{\sigma}_i - \sigma\right|}{\sum\limits_{i=1}^N\left|\sigma_i\right|}$ & Evaluates the relative error between the true value and prediction \cite{9128764}. \\
        &  &\\
        Image Correlation Coefficient (ICC) & $\dfrac{\sum\limits_{i=1}^N\left( \hat{\sigma}_i- \mu_{\hat{\boldsymbol{\sigma}}}\right)\left( \sigma_i-\mu_{ \boldsymbol{\sigma}}\right)}{\sqrt{\sum\limits_{i=1}^N\left( \hat{\sigma}_i-\mu_{ \hat{\boldsymbol{\sigma}}}\right)^2} \sqrt{\sum\limits_{i=1}^N\left( \sigma_i-\mu_{ \boldsymbol{\sigma}}\right)^2}}$ & Measures the similarity between the true value and prediction\cite{9128764, yang2023eit}.\\
        & &\\
        Dice Coefficient (DC) & $\dfrac{2|X \cap Y|}{|X|+|Y|}$& Tests the accuracy of the results. It provides a ratio of pixels correctly predicted to the total number of pixels—the closer to 1, the better \cite{guo2022transformer}. For our experiments, we round the pixel values to 2 decimal places before evaluation. \\
        &  &\\
        Relative $L^2$ Error (RLE)  &$\dfrac{\| \hat{\boldsymbol{\sigma}}- \boldsymbol{\sigma}\|_2}{\| \boldsymbol{\sigma}\|_2} = \dfrac{\left(\sum\limits_{i=1}^N\left|\hat{\sigma}_i - \sigma\right|^2\right)^{1 / 2}}{\left(\sum\limits_{i=1}^N\left|\sigma_i\right|^2\right)^{1 / 2}}$ & Measures the relative difference between the truth and the prediction. The closer to $0$, the better. \cite{guo2022transformer, yang2023eit} \\
        & &\\
        Root Mean Squared Error (RMSE) & $\sqrt{\dfrac{1}{N} \sum\limits_{i=1}^N\left(\sigma_i-\hat{\sigma}_i\right)^2}$ &  Evaluates the average magnitude of the differences between the truth and the prediction. \cite{yang2023eit}\\
        &  &\\
        Mean Absolute Error (MAE) & $\dfrac{1}{N} \sum\limits_{i=1}^N |\sigma_i-\hat{\sigma}_i|$ & Evaluates the average magnitude of the differences between the truth and the prediction\cite{yang2023eit} \\ 
        \bottomrule
        \end{tabular}
	\caption{Description of various performance metrics.} 
    \label{tab:perf_metric}
\end{table}

\subsection{Results and discussions}


Tables \ref{tab:performance-pwc} and \ref{tab:performance-text} present quantitative values for the performance metrics of various EIT reconstruction methods, in the presence of different noise levels, $\delta = 0\%$, $\delta = 1\%$ and, $\delta = 5\%$. The considered performance metrics are described in Table \ref{tab:perf_metric}. Understanding the results requires considering the behaviour of these metrics: 
For RIE, RMSE, MAE, and RLE, lower values indicate better performance and the objective is to minimise them; for DC and ICC, values closer to 1 indicate better performance, and the goal is to maximise them.
Below, we examine the results in each table more closely.
\begin{table}[!ht]\small
    \centering
    \begin{subtable}[htp]{\textwidth}
        \centering
        \begin{tabular}{rcccccc}\toprule
         & RIE & ICC & DC & RMSE & MAE & RLE \\  \midrule
        Sparsity & $0.03844$ & $0.02159$ & \cellcolor{gray!20}$0.79134$ & $0.09989$ & $0.10360$ & $0.03904$ \\
        D-bar & $0.09784$ & $0.01486$ & $0.08581$ &	$0.15515$ &	$0.16123$ &	$0.09928$ \\ \hdashline
        Deep D-bar & $0.03677$ & $0.02627$ & $0.45121$ & $0.09957$ & $0.10269$ & $0.03721$\\
        DDSM & $0.03450$ & $0.02690$ & $0.48590$ & $0.08793$ & $0.09075$ & $0.03494$\\
        FC-UNet & \cellcolor{gray!20}$0.01863$ & \cellcolor{gray!20}$0.02954$ & $0.76004$ & \cellcolor{gray!20}$0.06405$ & \cellcolor{gray!20}$0.06615$ & \cellcolor{gray!20}$0.01890$\\
        CNN LeNet & $0.04951$ & $0.02509$ & $0.18129$ & $0.08579$ & $0.08856$ & $0.05011$\\  \bottomrule
        \end{tabular}
        \caption{$\delta = 0\%$}
        \label{tab:pwc-delta-0}
    \end{subtable}
    \newline
    \vspace*{0.25cm}
    \begin{subtable}[htp]{\textwidth}
        \centering
        \begin{tabular}{rcccccc}\toprule
         & RIE & ICC & DC & RMSE & MAE & RLE \\  \midrule
        Sparsity & $0.03835$ & $0.02162$ & \cellcolor{gray!20}$0.79263$ & $0.09982$ & $0.10353$ & $0.03894$ \\ 
        D-bar & $0.08756$ & $0.01429$ & $0.08125$ & $0.14254$ & $0.14798$ & $0.08889$ \\ \hdashline
        Deep D-bar & $0.02738$ & $0.02762$ & $0.75264$ & $0.08477$ & $0.08751$ & $0.02774$\\
        DDSM & $0.03581$ & $0.02705$ & $0.46511$ & $0.09047$ & $0.09342$ & $0.03630$\\
        FC-UNet & \cellcolor{gray!20}$0.02159$ & \cellcolor{gray!20}$0.02929$ & $0.72974$ & \cellcolor{gray!20}$0.07170$ & \cellcolor{gray!20}$0.07409$ & \cellcolor{gray!20}$0.02194$\\
        CNN LeNet & $0.05905$ & $0.02499$ & $0.15198$ & $0.09884$ & $0.10215$ & $0.05988$\\  \bottomrule
        \end{tabular}
        \caption{$\delta = 1\%$}
        \label{tab:pwc-delta-0.01}
    \end{subtable}
    \newline
    \vspace*{0.25cm}
    \begin{subtable}[htp]{\textwidth}
        \centering
        \begin{tabular}{rcccccc}\toprule
         & RIE & ICC & DC & RMSE & MAE & RLE \\  \midrule
        Sparsity & \cellcolor{gray!20}$0.03952$ & $0.02159$ & \cellcolor{gray!20} $0.78729$ & \cellcolor{gray!20}$0.10272$ & \cellcolor{gray!20}$0.10658$ & \cellcolor{gray!20}$0.04015$\\
        D-bar & $0.08585$ & $0.01349$ & $0.06841$ & $0.13870$ & $0.14377$ & $0.08713$ \\ \hdashline
        Deep D-bar & $0.05563$ & $0.02272$ & $0.51711$ & $0.13523$ & $0.13954$ & $0.05648$\\
        DDSM & $0.04833$ & $0.02412$ & $0.38292$ & $0.11310$ & $0.11704$ & $0.04917$\\
        FC-UNet & $0.04332$ & \cellcolor{gray!20}$0.02672$ & $0.28293$ & $0.11519$ & $0.11923$ & $0.04415$\\
        CNN LeNet & $0.13901$ & $0.02312$ & $0.04568$ & $0.21138$ & $0.21876$ & $0.14155$\\  \bottomrule
        \end{tabular}
        \caption{$\delta = 5\%$}
        \label{tab:pwc-delta-0.05}
    \end{subtable}
	\caption{The performance of various methods trained and tested with piece-wise constant data at different noise levels. The neural networks used were trained with noiseless measurements for the deep learning-based methods.}
    \label{tab:performance-pwc}
\end{table}

\begin{figure}[!ht]
     \centering
     \begin{subfigure}[b]{0.560\textwidth}
         \centering
         \includegraphics[width=\textwidth]{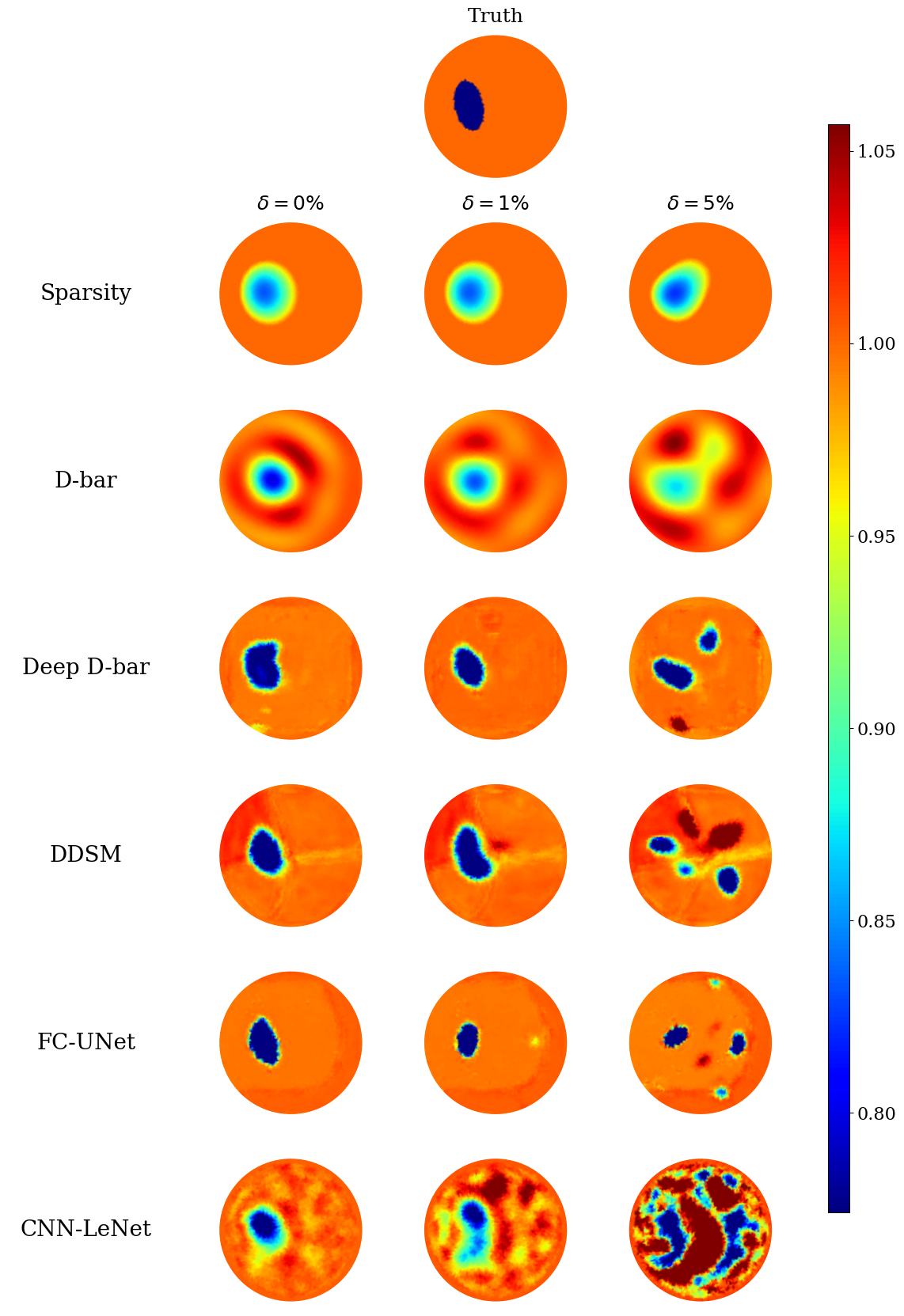}
         \caption{Sample 1.}
         \label{fig:pwc-samp1}
     \end{subfigure}
     \hfill
     \begin{subfigure}[b]{0.428\textwidth}
         \centering
         \includegraphics[width=\textwidth]{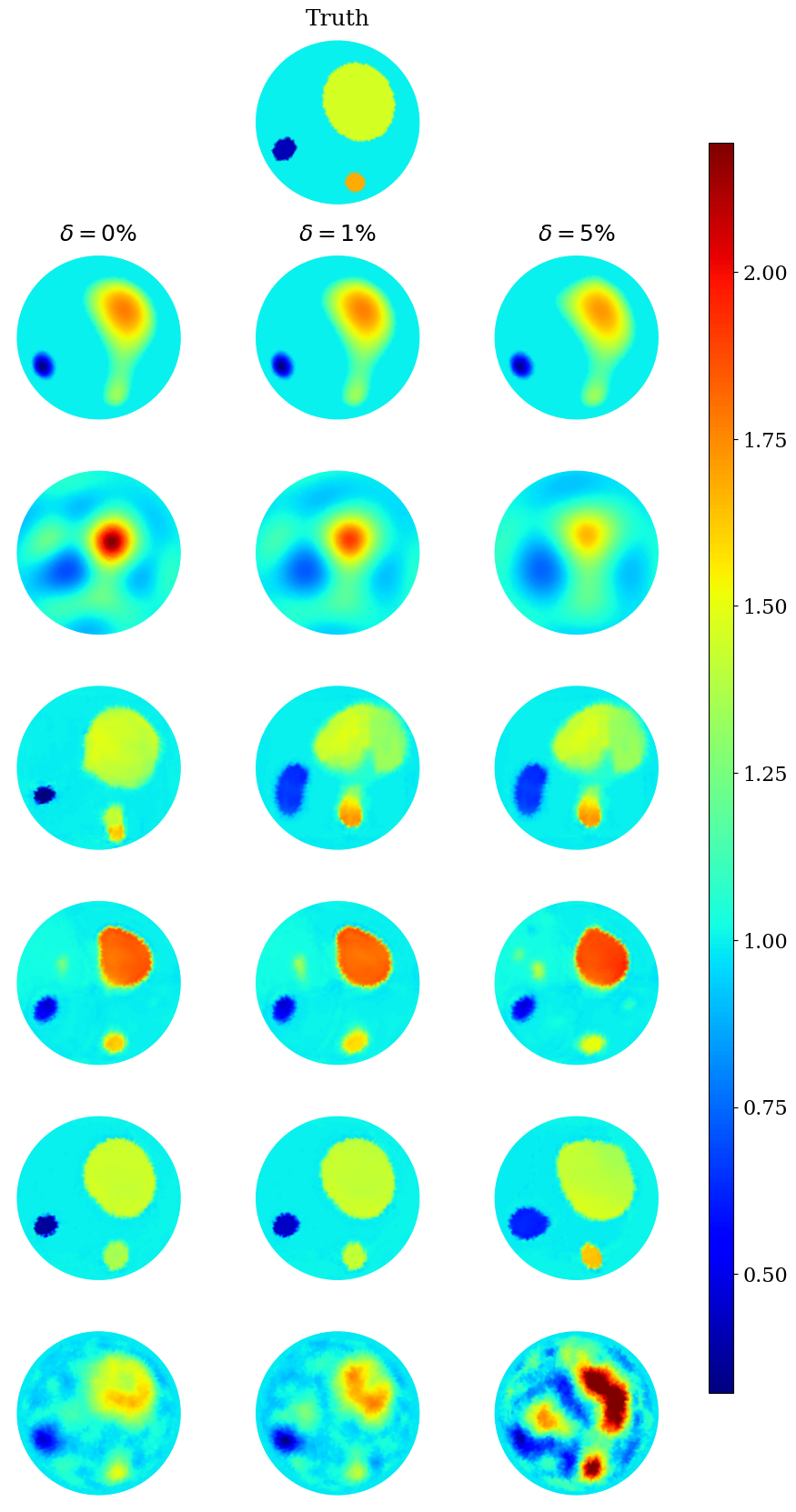}
         \caption{Sample 2.}
         \label{fig:pwc-samp2}
     \end{subfigure}
     \caption{Effects of noise on two piecewise constant samples by various reconstruction methods.}
     \label{fig:pwc-samples}
\end{figure}

\subsubsection{Piece-wise constant conductivities}
In the noiseless scenario as depicted in Table \ref{tab:pwc-delta-0}, FC-UNet shows the best performance across all metrics, with notably low RIE, RMSE, MAE, and RLE. It also achieves a high DC and ICC, indicating robustness and accuracy in image reconstruction. 
The DDSM also performs well, particularly regarding RIE, RMSE, MAE, and RLE. The Deep D-bar method exhibits competitive results, although slightly inferior to FC-UNet.
Both Sparsity and D-bar methods show weaker performance compared to the deep learning-based methods.
The CNN-LeNet method generally has the worst performance metrics, indicating less accurate image reconstruction.

Under increased noise of $\delta = 1\%$, the relative performance of the methods remains consistent, with FC-UNet still demonstrating strong performance. Also, the Deep D-bar performs exceptionally well in this case, particularly in terms of RIE, RMSE, MAE, and RLE. The DDSM also exhibits robust performance under this noise level, while the CNN LeNet method continues to have the highest values for most metrics, indicating challenges in handling noise. In contrast, the analytic-based methods of Sparsity and D-bar show particular robustness to the added noise, evidenced by the unnoticeable change in the performance metrics.

At a higher noise level $\delta = 5\%$, the inverse problem becomes more challenging due to the severe ill-posed nature; and in the learned context, since the neural networks are trained on noiseless data, which differ markedly from the noisy data, the setting may be viewed as an out-of-distribution robustness test. Here, the sparsity method comes on top across most metrics, having almost maintained constant performance.  However, the FC-UNet continues to maintain the best performance in terms of  ICC, emphasising its robustness in noisy conditions.
Deep D-bar and DDSM display competitive results, indicating resilience to increased noise.
The D-bar methods exhibit slightly weaker performance, especially in terms of RIE, RMSE, and MAE. In contrast, the CNN LeNet method continues to have the highest values for most metrics, suggesting difficulty in coping with substantial noise. 

Overall, these results illustrate the varying performance of different EIT methods under different noise levels. The deep learning-based methods, particularly FC-UNet, exhibit good performance across low noise levels. In contrast, the sparsity method shows proof of consistent robustness across higher noise levels, indicating their effectiveness in reconstructing EIT images, even in the presence of noise. Visual results across all the noise levels are shown for two test samples in Figure \ref{fig:pwc-samples}.

\subsubsection{Textured inclusions scenario}
In the noiseless scenario depicted in Table \ref{tab:text-delta-0}, The best-performing method based on RIE, ICC, RMSE, MAE, and RLE is FC-UNet, with the best values across these metrics. The sparsity method and DDSM also perform well, being the first runners-up in these metrics, particularly for DC; the sparsity method achieves the highest values, indicating good performance, with DDSM as the first runner-up. The worst-performing method across all metrics in this scenario is "D-bar."
\begin{table}[!ht]\small
    \centering
    \begin{subtable}[htp]{\textwidth}
        \centering
        \begin{tabular}{rcccccc}\toprule
         & RIE & ICC & DC & RMSE & MAE & RLE \\  \midrule
        Sparsity & $0.03869$ & $0.01968$ & \cellcolor{gray!20}$0.79695$ & $0.10473$ & $0.10864$ & $0.03949$ \\ 
        D-bar & $0.08856$ & $0.01473$ & $0.09956$ & $0.14597$ & $0.15257$ & $0.09063$ \\ \hdashline
        Deep D-bar & $0.03677$ & $0.02627$ & $0.45121$ & $0.09957$ & $0.10269$ & $0.03721$\\
        DDSM & $0.03559$ & $0.02360$ & $0.42812$ & $0.08930$ & $0.09282$ & $0.03641$ \\
        FC-UNet & \cellcolor{gray!20}$0.02781$ &	\cellcolor{gray!20}$0.02639$ & $0.45464$ & \cellcolor{gray!20}$0.07379$ & \cellcolor{gray!20}$0.07679$ & \cellcolor{gray!20}$0.02850$ \\
        CNN LeNet & $0.04930$ &	$0.02308$ & $0.18749$ & $0.09010$ & $0.09357$ & $0.05034$ \\  \bottomrule
        \end{tabular}
        \caption{$\delta = 0\%$}
        \label{tab:text-delta-0}
    \end{subtable}
    \newline
    \vspace*{0.25cm}
    \begin{subtable}[htp!]{\textwidth}
        \centering
        \begin{tabular}{rcccccc}\toprule
         & RIE & ICC & DC & RMSE & MAE & RLE \\  \midrule
        Sparsity & $0.03871$ & $0.01961$ & \cellcolor{gray!20}$0.79497$ & $0.10455$ & $0.10846$ & $0.03951$ \\ 
        D-bar & $0.07982$ & $0.01381$ & $0.09294$ & $0.13634$ & $0.14231$ & $0.08168$ \\ \hdashline
        Deep D-bar & \cellcolor{gray!20}$0.02738$ & \cellcolor{gray!20}$0.02762$ & $0.75264$ & $0.08477$ & $0.08751$ & \cellcolor{gray!20}$0.02774$\\
        DDSM & $0.03663$ & $0.02325$ & $0.43803$ & $0.09261$ & $0.09626$ & $0.03750$ \\
        FC-UNet & $0.02980$ &	$0.02592$ & $0.42355$ & \cellcolor{gray!20}$0.07957$ & \cellcolor{gray!20}$0.08280$ & $0.03055$ \\
        CNN LeNet & $0.06301$ &	$0.02253$ & $0.12531$ & $0.10709$ & $0.11112$ & $0.06426$ \\  \bottomrule
        \end{tabular}
        \caption{$\delta = 1\%$}
        \label{tab:text-delta-0.01}
    \end{subtable}
    \newline
    \vspace*{0.25cm}
    \begin{subtable}[!ht]{\textwidth}
        \centering
        \begin{tabular}{rcccccc}\toprule
         & RIE & ICC & DC & RMSE & MAE & RLE \\  \midrule
        Sparsity & \cellcolor{gray!20}$0.03975$ & $0.01961$ & \cellcolor{gray!20}$0.79157$ & \cellcolor{gray!20}$0.10766$ & \cellcolor{gray!20}$0.11177$ & \cellcolor{gray!20}$0.04061$\\
        D-bar & $0.08015$ & $0.01265$ & $0.07503$ & $0.13590$ & $0.14161$ & $0.08193$ \\ \hdashline
        Deep D-bar & $0.05563$ & \cellcolor{gray!20}$0.02272$ & $0.51711$ & $0.13523$ & $0.13954$ & $0.05648$\\
        DDSM & $0.04775$ & $0.02091$ & $0.38445$ & $0.11451$ &	$0.11883$ &	$0.04882$ \\
        FC-UNet & $0.05195$ &	$0.02269$ & $0.28514$ & $0.12528$ & $0.12999$ & $0.05312$ \\
        CNN LeNet & $0.17301$ &	$0.01907$ & $0.03654$ & $0.25557$ & $0.26481$ & $0.17637$ \\  \bottomrule
        \end{tabular}
        \caption{$\delta = 5\%$}
        \label{tab:text-delta-0.05}
    \end{subtable}
	\caption{The performance of various methods trained and tested with textured data at different noise levels. The neural networks used were trained with noiseless measurements for the deep learning-based methods.}
    \label{tab:performance-text}
\end{table}
With a bit of noise of $1 \%$ added, the Deep D-bar surprisingly stands out as the best-performing for most of the considered metrics. The FC-UNet closely follows it. The sparsity-based method continues to lead in DC. Like the noiseless scenario, D-bar remains one of the less effective methods across all metrics. This is depicted in Table \ref{tab:text-delta-0.01}.

For higher noise levels in Table \ref{tab:text-delta-0.05}, the sparsity-based methods once again excel in all metrics but for the ICC, making it the best-performing method. The DDSM and FC-UNet closely follow in most of these metrics, while the Deep D-bar continues to perform best in ICC. The CNN LeNet consistently performs the poorest across all metrics and noise levels, especially in this high-noise scenario.

\begin{figure}[!ht]
     \centering
     \begin{subfigure}[b]{0.560\textwidth}
         \centering
         \includegraphics[width=\textwidth]{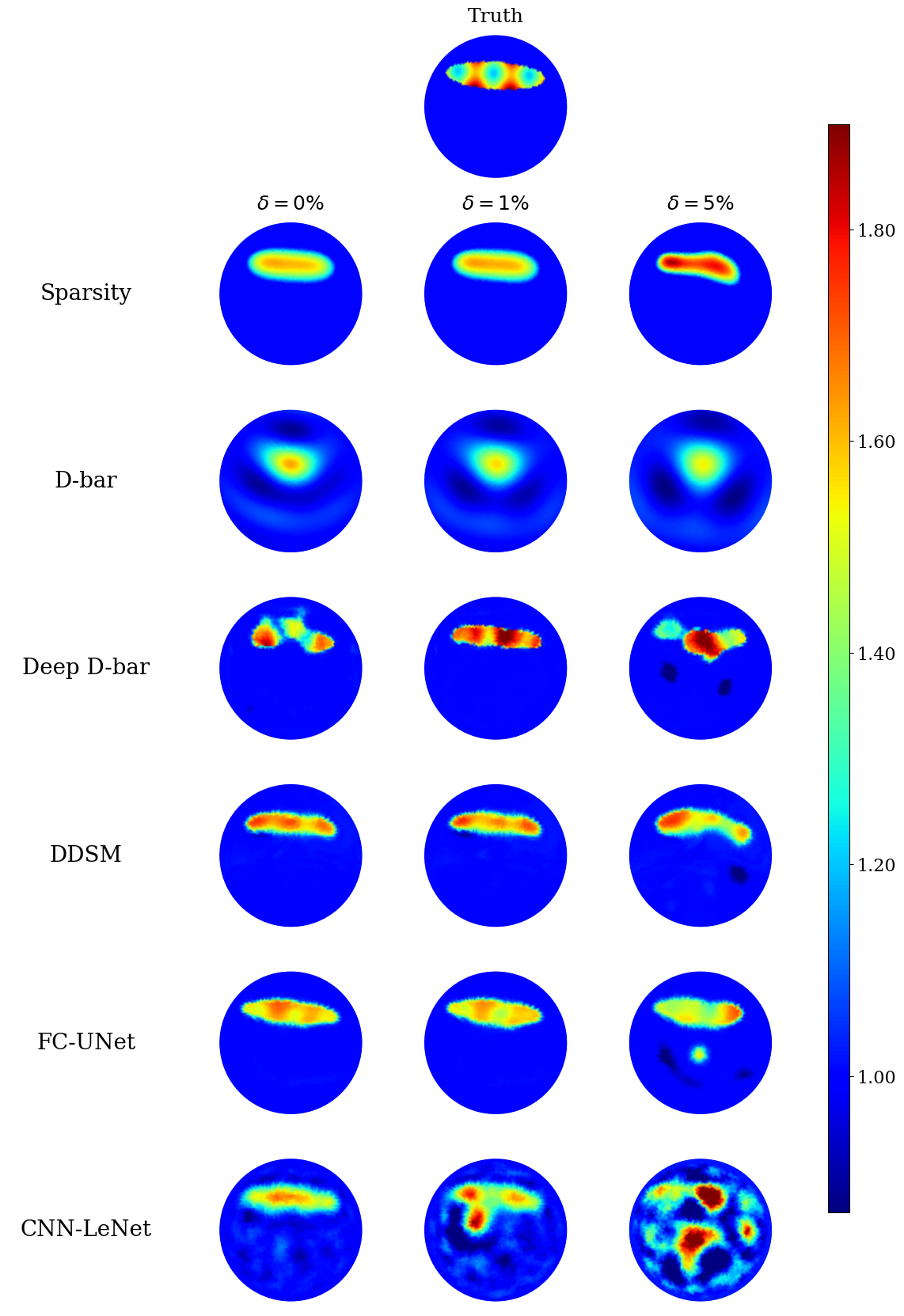}
         \caption{Sample 1.}
         \label{fig:tex-samp1}
     \end{subfigure}
     \hfill
     \begin{subfigure}[b]{0.428\textwidth}
         \centering
         \includegraphics[width=\textwidth]{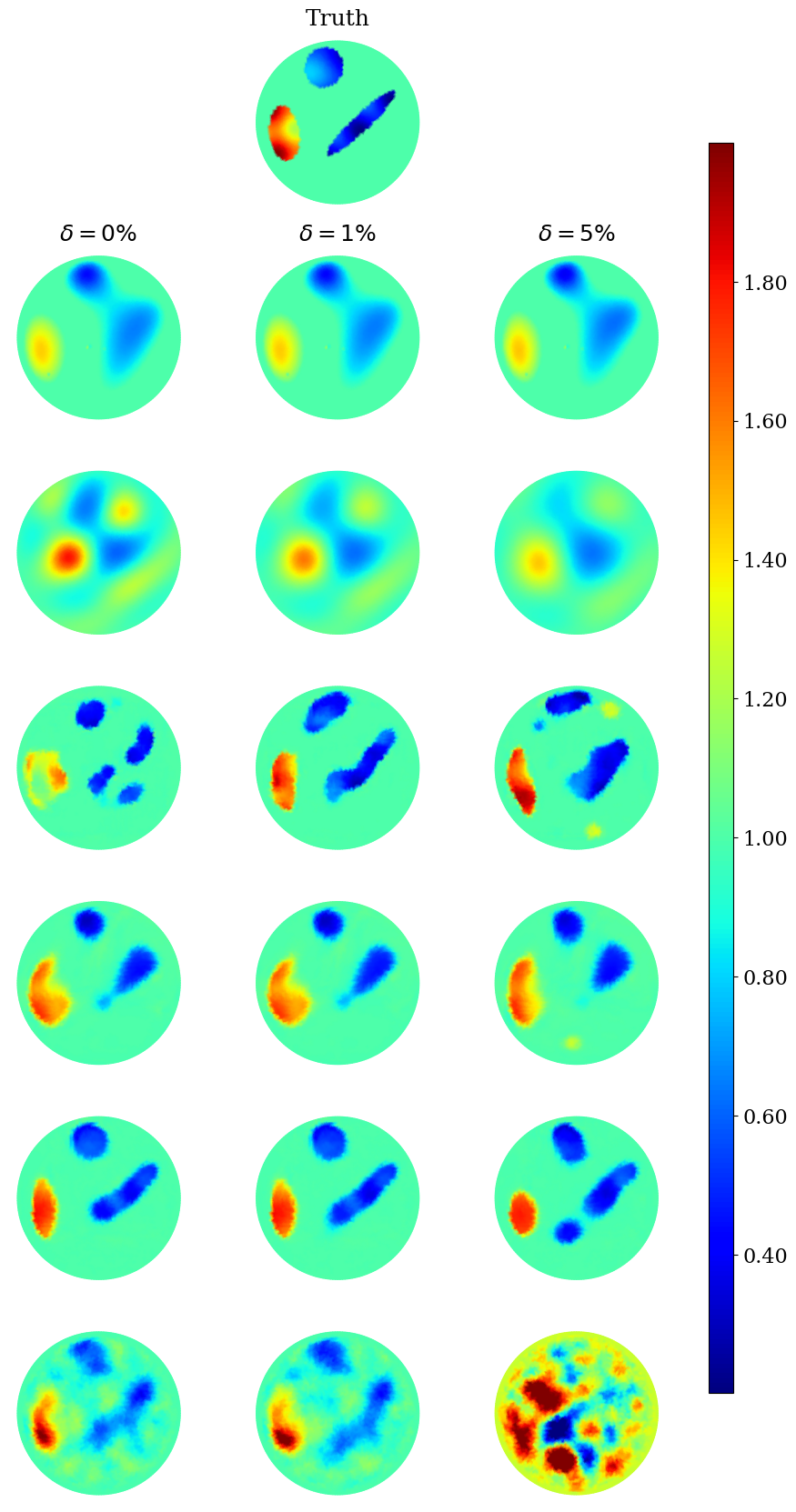}
         \caption{Sample 2.}
         \label{fig:tex-samp2}
     \end{subfigure}
     \caption{Effects of noise on two textured samples by various reconstruction methods.}
     \label{fig:tex-samples}
\end{figure}

In summary, the best-performing method varies depending on the specific performance metric and noise level. Sparsity consistently demonstrates robust performance in both noiseless and noisy scenarios, while the D-bar is generally less effective. However, in terms of computational expense, the sparsity method is more expensive. The Deep D-bar, FC-UNet, and DDSM often serve as strong contenders, shifting their rankings across noise scenarios and metrics. Meanwhile, CNN LeNet consistently performs the poorest, particularly in high-noise scenarios ($\sigma = 5\%$).
Figure \ref{fig:tex-samples} depicts this for two test examples.

\begin{figure}[!ht]
     \centering
     \begin{subfigure}[b]{\textwidth}
         \centering
         \includegraphics[width=\textwidth]{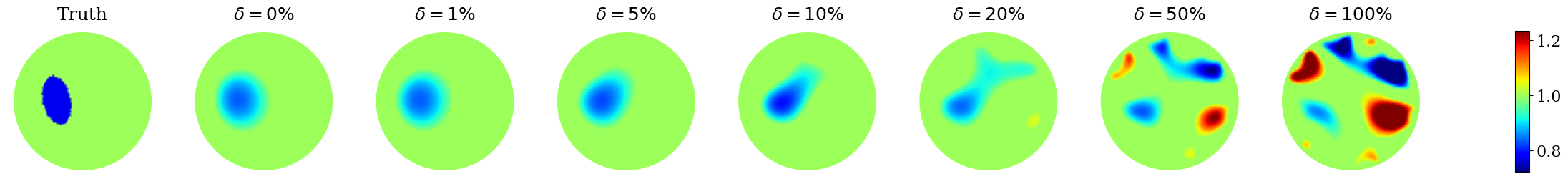}
         \caption{Sample 1.}
     \end{subfigure}\\
     \begin{subfigure}[b]{\textwidth}
         \centering
         \includegraphics[width=\textwidth]{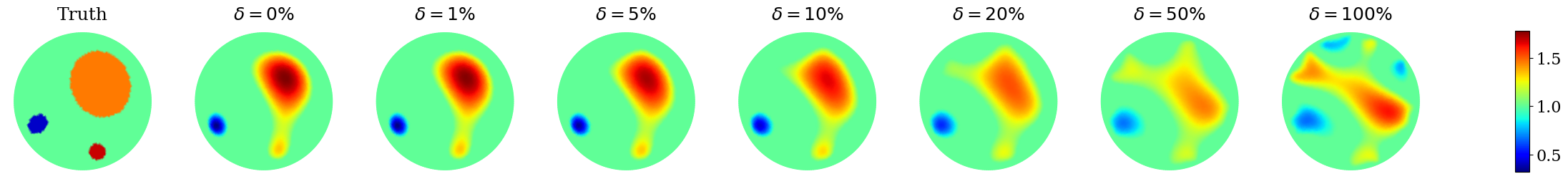}
         \caption{Sample 2.}
     \end{subfigure}
     \caption{Effects of additional noise on two piecewise samples by the sparsity method.}
     \label{fig:pwc-bignoise-samples}
\end{figure}

\begin{figure}[!ht]
     \centering
     \begin{subfigure}[b]{0.5\textwidth}
         \centering
         \includegraphics[width=0.95\textwidth]{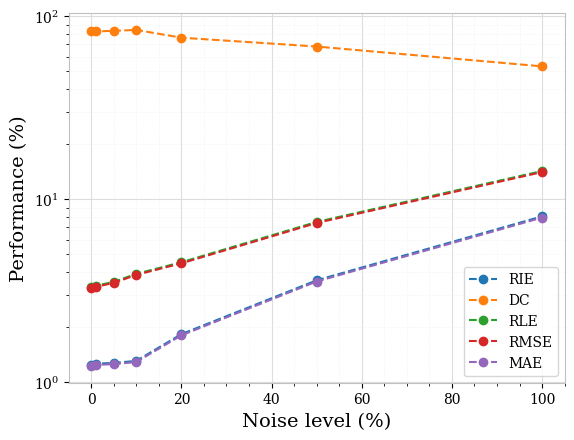}
         \caption{Sample 1.}
     \end{subfigure}\hfill
     \begin{subfigure}[b]{0.5\textwidth}
         \centering
         \includegraphics[width=0.95\textwidth]{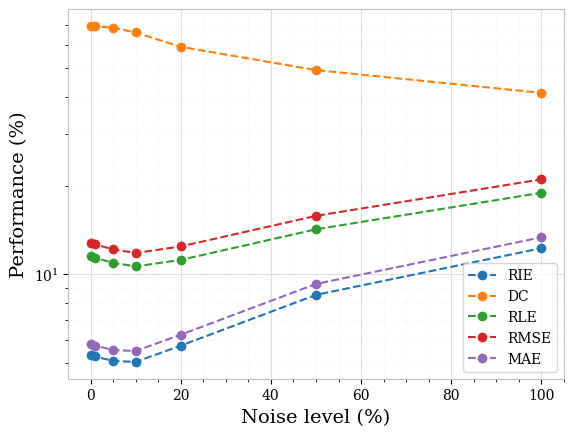}
         \caption{Sample 2.}
     \end{subfigure}
     \caption{Performance variation with noise for two piecewise samples by the sparsity method.}
     \label{fig:pwc-bignoise-variation}
\end{figure}

\begin{figure}[!ht]
     \centering
     \begin{subfigure}[b]{\textwidth}
         \centering
         \includegraphics[width=\textwidth]{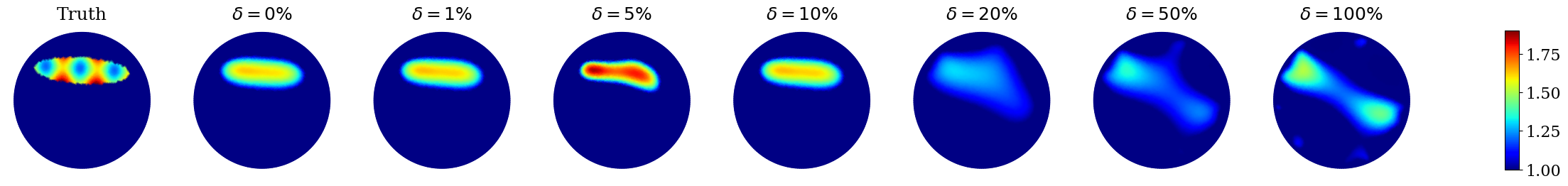}
         \caption{Sample 1.}
     \end{subfigure}\\
     \begin{subfigure}[b]{\textwidth}
         \centering
         \includegraphics[width=\textwidth]{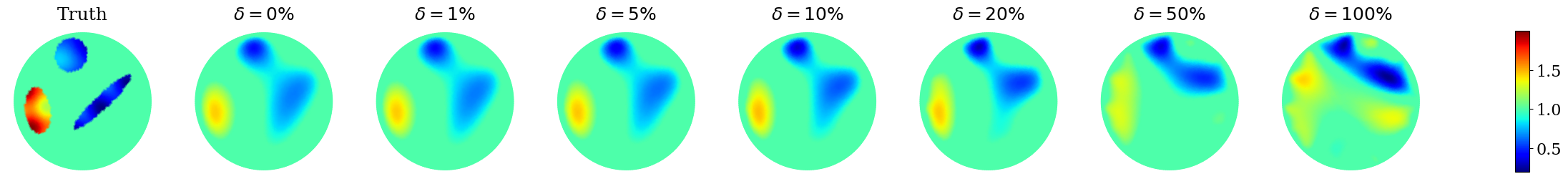}
         \caption{Sample 2.}
     \end{subfigure}
     \caption{Effects of additional noise on two textured samples by the sparsity method.}
     \label{fig:tex-bignoise-samples}
\end{figure}

\begin{figure}[!ht]
     \centering
     \begin{subfigure}[b]{0.5\textwidth}
         \centering
         \includegraphics[width=0.95\textwidth]{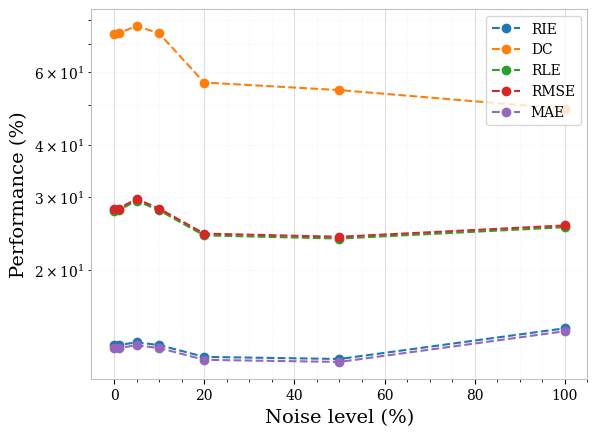}
         \caption{Sample 1.}
     \end{subfigure}\hfill
     \begin{subfigure}[b]{0.5\textwidth}
         \centering
         \includegraphics[width=0.95\textwidth]{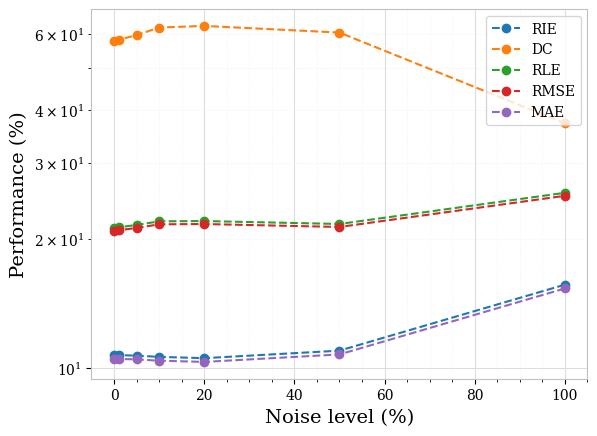}
         \caption{Sample 2.}
     \end{subfigure}
     \caption{Performance variation with noise for two textured samples by the sparsity method.}
     \label{fig:tex-bignoise-variation}
\end{figure}

Furthermore, for both piecewise constant and textured phantoms, the sparsity-based method consistently performed well for noisy scenarios. This consistently good performance of the sparsity concept in detecting and locating inclusions even for higher noise levels is most remarkable. The error metrics are almost constant over noise levels up to $5\%$. Hence, as a side result, we did check the limits of the sparsity concept for very high noise levels, which not surprisingly showed a sharp decrease in the reconstruction accuracy for very high noise levels. We show this in Figure \ref{fig:pwc-bignoise-samples}, once again for the two piecewise constant samples initially displayed in Figure \ref{fig:pwc-samples}. The respective performances, all metrics considered, for these two samples are equally shown in Figure \ref{fig:pwc-bignoise-variation} (ICC is not plotted for the sake of visibility since its values are smallest). Figures \ref{fig:tex-bignoise-samples} and \ref{fig:tex-bignoise-variation} show the corresponding plots for the textured samples initially displayed in Figure \ref{fig:tex-samples}.

\section{Conclusion and future directions}\label{sec:discussion}
In summary, this review has comprehensively examined numerical methods for addressing the EIT inverse problem. EIT, a versatile imaging technique with applications in various fields, presents a highly challenging task of reconstructing internal conductivity distributions from boundary measurements.
We explored the interplay between modern deep learning-based approaches and traditional analytic methods for solving the EIT inverse problem. Four advanced deep learning algorithms were rigorously assessed, including the deep D-bar method, deep direct sampling method, fully connected U-net, and convolutional neural networks. Additionally, two analytic-based methods, incorporating mathematical formulations and regularisation techniques, were examined regarding their efficacy and limitations.
Our evaluation involved a comprehensive array of numerical experiments encompassing diverse scenarios that mimic real-world complexities. Multiple performance metrics were employed to shed insights into the methods' capabilities to capture essential features and delineate complex conductivity patterns.

The first evaluation was based on piecewise constant conductivities. The clear winners of this series of tests are the analytic sparsity-based reconstruction and the learned FC-UNet. Both perform best, with slight variations depending on the noise level. This is not surprising for learned methods, which adapt well to this particular set of test data. However, the excellent performance of sparsity methods, which can identify and locate piecewise constant inclusions correctly, is most remarkable.

A noteworthy aspect of this study was the introduction of variable conductivity scenarios, mimicking textured inclusions and departing from uniform conductivity assumptions. This enabled us to assess how each method responds to varying conductivity, shedding light on their robustness and adaptability. Here, the D-bar with learned post-processing achieves competitive results. The winning algorithm alternates between sparsity, Deep D-bar and FC-UNet. The good performance of the sparsity concepts is somewhat surprising for these textured test samples. However, none of the proposed methods was able to reconstruct the textures reliably for higher noise levels. That is, the quality of the reconstruction was mainly measured in terms of how well the inclusions were located - which gives a particular advantage to sparsity concepts.

These results naturally raise questions about the numerical results presented in several existing EIT studies, where learned methods were only compared with sub-optimal analytic methods. Our findings clearly indicate that at least within the restricted scope of the present study, optimised analytical methods can reach a comparable or even superior accuracy. Of course, one should note that after training, learned methods are much more efficient and provide a preferred option for real-time imaging.

In conclusion, this review contributes to a deeper understanding of the available solutions for the EIT inverse problem, highlighting the role of deep learning and analytic-based methods in advancing the field.

\section*{Acknowledgements}
    D.N.T. acknowledges the financial support of this research work within the Research Unit 3022 "Ultrasonic Monitoring of Fiber Metal Laminates Using Integrated Sensors" by the German Research Foundation (Deutsche Forschungsgemeinschaft (DFG)) under grant number LO1436/12-1 and project number 418311604.
    
    \noindent J.N. acknowledges the financial support from the program of China Scholarships Council (No. 202006270155).
    
    \noindent A.H. acknowledges support by the Research Council of Finland: Academy Research Fellow (Project No. 338408) and the Centre of Excellence of Inverse Modelling and Imaging project (Project No. 353093).
    
    \noindent B.J. acknowledges the support by a start-up fund and Direct Grant of Research, both from The Chinese University of Hong Kong, Hong Kong General Research Fund (Project No. 14306423) and UK Engineering and Physical Research Council (EP/V026259/1).
    
    \noindent P.M. acknowledges the financial support from the DFG project number 281474342:  Graduiertenkolleg RTG 2224 Parameter Identification - Analysis, Algorithms, Applications.

\bibliographystyle{abbrv}
\bibliography{eit-manuscript}

\end{document}